\documentclass[runningheads]{llncs}
\usepackage{graphicx}

\usepackage[svgnames ]{xcolor}
\newcommand{\bestcolor}[1]{\colorbox{Gold}{#1}}
\newcommand{\secondbestcolor}[1]{\colorbox{Silver}{#1}}

\usepackage{lipsum}
\usepackage{tikz}

\usepackage{comment}
\usepackage{amsmath,amssymb,amsfonts} % define this before the line numbering.
\usepackage{color}
\usepackage[pagebackref,breaklinks,colorlinks]{hyperref}
\usepackage[accsupp]{axessibility}  % Improves PDF readability for those with disabilities.
\usepackage{mathtools}
\usepackage{multirow}
\usepackage{booktabs}
\usepackage{float}
\usepackage{caption}
\usepackage{subcaption}
\usepackage{threeparttable}

\usepackage[capitalize]{cleveref}
\crefname{section}{Sec.}{Secs.}
\Crefname{section}{Section}{Sections}
\Crefname{table}{Table}{Tables}
\crefname{table}{Tab.}{Tabs.}
\crefname{equation}{Eq.}{Eqs.}

\begin{document}
% \linenumbers
\pagestyle{headings}
\mainmatter
\def\ECCVSubNumber{979}  % Insert your submission number here

\title{Sobolev Training for Implicit Neural Representations with Approximated Image Derivatives} % Replace with your title

\renewcommand{\thefootnote}{\fnsymbol{footnote}}
\footnotetext{This work is done by the first four authors as interns at Megvii Research.}

\titlerunning{Sobolev Training for INRs}
\author{Wentao Yuan$^{1,2}$
\,\,
Qingtian Zhu$^1$
\,\,
Xiangyue Liu$^1$ 
\,\,
Yikang Ding$^1$
\,\, \\
Haotian Zhang$^1$\thanks{Corresponding author (zhanghaotian@megvii.com).}
\,\,
Chi Zhang$^1$ 
}
\institute{$^1$Megvii Research \quad $^2$Peking University 
}
\authorrunning{W. Yuan et al.}

%******************
\maketitle

\begin{abstract}

Recently, Implicit Neural Representations (INRs) parameterized by neural networks have emerged as a powerful and promising tool to represent different kinds of signals due to its continuous, differentiable properties, showing superiorities to classical discretized representations. However, the training of neural networks for INRs only utilizes input-output pairs, and the derivatives of the target output with respect to the input, which can be accessed in some cases, are usually ignored. In this paper, we propose a training paradigm for INRs whose target output is image pixels, to encode image derivatives in addition to image values in the neural network.
Specifically, we use finite differences to approximate image derivatives.
We show how the training paradigm can be leveraged to solve typical INRs problems, i.e., image regression and inverse rendering, and demonstrate this training paradigm can improve the data-efficiency and generalization capabilities of INRs. The code of our method is available at \url{https://github.com/megvii-research/Sobolev_INRs}. 

\end{abstract}

\section{Introduction}

\begin{figure}[t]
    \centering
    \includegraphics[width=\linewidth]{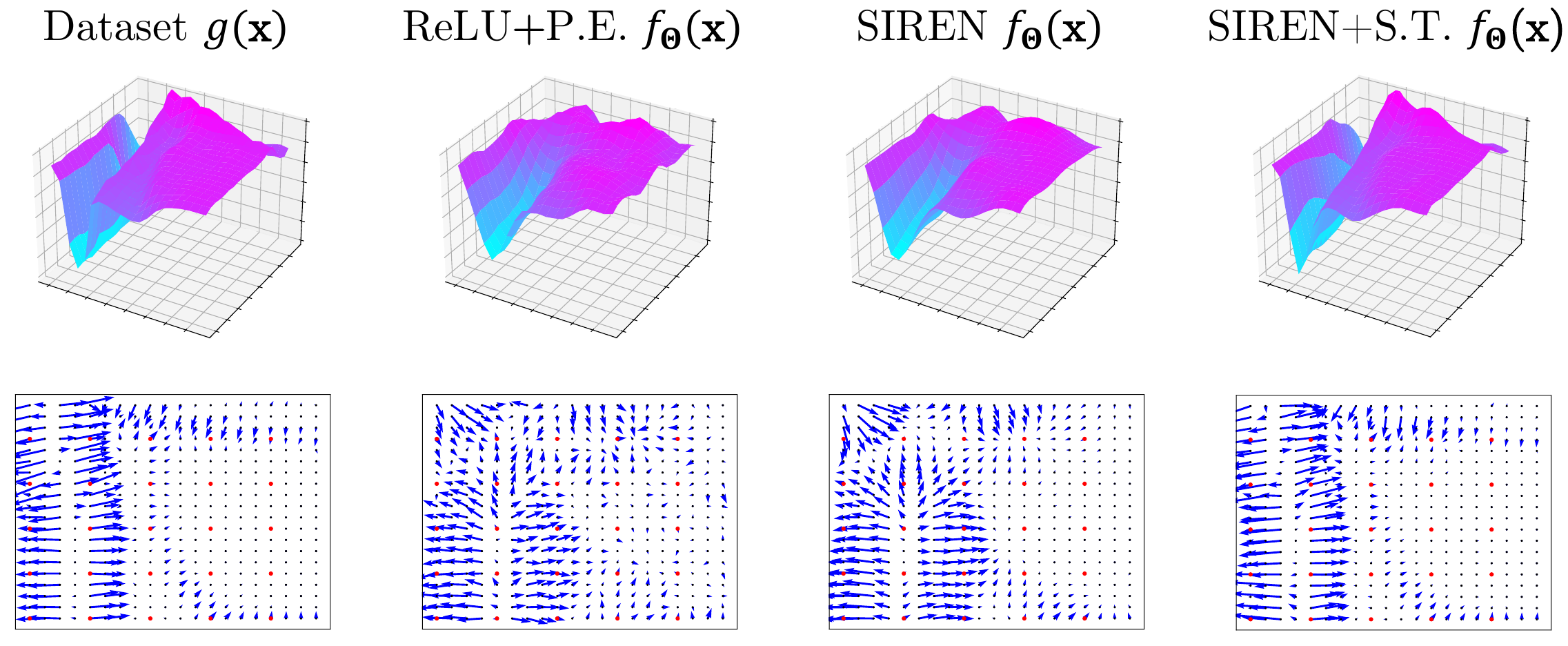}
    \caption{Visualized approximation results of different training schemes and different activation functions. P.E. stands for positional encoding~\cite{mildenhall2020nerf,tancik2020fourier}; S.T. stands for Sobolev training. Top: dataset $g$ (the target function to approximate) as well as approximated functions $f_{\boldsymbol{\Theta}}$ (functions parameterized by neural networks) represented by MLPs within an image patch of $20\times 20$ pixels. Bottom: vector fields of the first-order derivatives, where red points are training samples (in a total proportion of 6.25\%) and remaining black points are unseen samples. When activated by sine functions and trained with additional supervision on derivatives, the network demonstrates the best approximation of both function values and derivatives. }
    \label{fig:field}
\end{figure}

A promising recent direction in computer vision and computer graphics is encoding complex signals implicitly by aid of multi-layer perceptron (MLP) named Implicit Neural Representations (INRs)~\cite{park2019deepsdf,mescheder2019occupancy,sitzmann2019scene,niemeyer2020differentiable,mildenhall2020nerf,sitzmann2020implicit,sitzmann2021light,suhail2022light}.
The MLPs (more specifically, coordinate-based MLPs) take low-dimensional coordinates as inputs and are trained to output a representation of shape, density, and/or colors at each input location. 

Compared to classical alternatives, i.e., discrete grid-based representation, INRs offer two main benefits. (a) Due to the fact that MLPs are continuous functions, they are significantly more memory efficient than discrete grid-based representations and theoretically, signals parameterized by MLPs can be presented in arbitrary resolution. (b) INRs are naturally differentiable, so gradient descent (GD) can be applied to optimize MLPs using modern auto-differentiation tools, e.g., PyTorch~\cite{paszke2019pytorch}, TensorFlow~\cite{abadi2016tensorflow}, JAX~\cite{jax2018github}.

However, as will be discussed below, the training paradigm adopted by most of the INRs only utilizes input-output pairs, which brings about bad generalization capabilities on unseen coordinates.
In other words, if trained with low-resolution training data, MLPs will yield heavily degraded predictions when the target outputs are high-resolution. 
Based on this observation, an intuitive idea is that we can take advantage of the differentiability of INRs to improve the generalization capabilities. 
In this paper, we propose a training paradigm for INRs whose training data and the desired results are both images, with both supervision on values and derivatives enforced. 
Considering that the ground truth representation (the ``continuous" image) is agnostic, finite differences are applied to approximate numerical first-order derivatives on images.

Further, we study some important peripheral problems relevant to the proposed training paradigm, e.g., the choice of activation functions, the number of layers in MLPs and the choice of image filters.

We find that in practice, most of the recent INRs build on ReLU-based MLPs~\cite{mildenhall2020nerf,tancik2020fourier,chen2021learning} fail to represent well the derivatives of target outputs, even if the corresponding supervision is provided.
For this reason, we use periodic activation functions~\cite{Parascandolo2017TamingTW,sitzmann2020implicit}, i.e., sine function, to improve MLPs' convergence property of derivatives.
As is shown in \cref{fig:field}, INRs trained with traditional value-based schemes fail to explicitly constraint the behavior at unseen coordinates, leading to poor generalization both in the image field and the vector field of the first-order derivatives. Besides, with sine-activated MLPs trained with both value loss and derivative loss, INRs are able to generalize well even if the training samples only take a very limited proportion (6.25\% in \cref{fig:field}).

We show how the training paradigm can be adapted to specific problems, namely direct image regression~\cite{sitzmann2020implicit,tancik2020fourier} and indirect inverse rendering~\cite{mildenhall2020nerf,tancik2020fourier}. Experiments results demonstrate that our training paradigm can improve the data-efficiency and generalization capabilities of INRs. The main contributions are summarized below.
\begin{enumerate}
\setlength{\itemsep}{0pt}
\setlength{\parsep}{0pt}
\setlength{\parskip}{0pt}
    \item[-] We propose a training paradigm for INRs whose training data as well as target outputs are image pixels. With the paradigm, image values and image derivatives are encoded into the INRs. Concretely, we use finite differences to approximate the derivatives of the ground truth signal. To the best of our knowledge, supervising network derivatives in INRs is minimally explored.
    \item[-] We study the factor of activation functions in INRs when applying the proposed training paradigm. By experiments, We use sine functions for better convergence property of derivatives.
    \item[-] Experiment results on two typical INRs problems demonstrate that with the proposed paradigm we are able to train better INRs - representations that require fewer training samples and generalise better on unseen inputs.
\end{enumerate}

\section{Related Work}

\subsection{Implicit Neural Representations}
Implicit Neural Representations (INRs), which usually represent an object as a multi-layer perceptron model that maps low-dimensional coordinates to signal values, have drawn a lot of attention recently. Since this representation is continuous and can capture fine details of signals, it has been widely applied to novel view synthesis~\cite{mildenhall2020nerf,martin2021nerf,lin2021barf,park2021nerfies,yu2021pixelnerf,yu2021plenoctrees,tewari2021advances}, shape representation~\cite{chen2019learning,park2019deepsdf,michalkiewicz2019implicit,chabra2020deep,atzmon2020sal,gropp2020implicit,zhang2021ners,yariv2020multiview}, and multi-view 3D reconstruction~\cite{mescheder2019occupancy,oechsle2021unisurf,chen2021mvsnerf,zhang2021learning}. As a milestone, Mildenhall et al.~\cite{mildenhall2020nerf} propose a novel view synthesis method that learns an implicit representation for a specific 3D scene by using a set of multi-view calibrated images. Recent work from Park et al.~\cite{park2019deepsdf} puts forward to learn a Signed Distance Function (SDF) to represent the shape of objects, and different SDFs can be inferred with different input latent codes. However, the (approximated) ground truth derivative information of the target signal has rarely been utilized to train INRs.

\subsection{Derivative Supervision}

Derivative information for neural networks has also been exploited in some previous work. Hornik~\cite{hornik1991approximation} proves the universal approximation theorems for neural networks in Sobolev spaces - a vector space of functions equipped with a norm that is a combination of $\ell_p$-norms of the function together with its derivatives up to a given order. Hornik~\cite{hornik1991approximation} shows that neural networks with non-constant, bounded, continuous activation functions, with continuous derivatives up to order \textit{K} are universal approximators in the Sobolev spaces of order \textit{K}.
Further, Czarnecki et al.~\cite{czarnecki2017sobolev} prove ReLU-based MLPs are universal approximators for function values and first-order derivatives theoretically and propose Sobolev training for neural networks. However, only the scenarios with analytical derivatives are covered and the ReLU-based MLPs show relatively poor convergence properties in practice.
Gropp et al.~\cite{gropp2020implicit} propose an \textit{Eikonal term} which is a first-order derivative related penalty term to regularize INRs. The Eikonal term is different from our explicit derivative supervision with ground truth derivatives approximation. Sitzmann et al.~\cite{sitzmann2020implicit} 
verify that INRs with periodic activation functions can fit derivatives robustly, while the supervision on both values and derivatives and the property of Sobolev training in INRs remain unexplored.

\section{Methodology}
\subsection{Formulation}\label{sec:formulation}

\begin{figure}[t]
    \centering
    \includegraphics[width=0.80\linewidth]{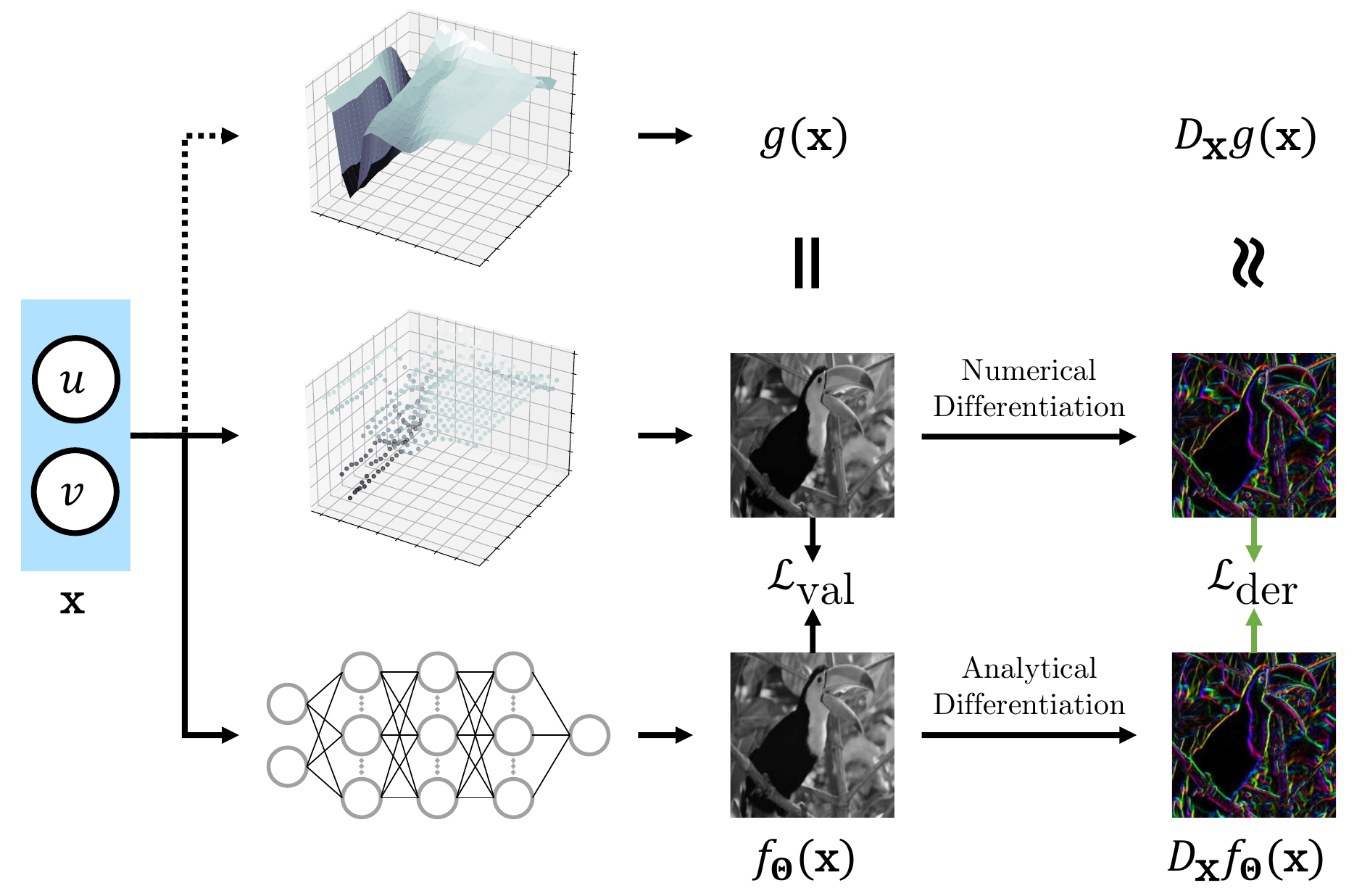}
    \caption{An overview of the proposed training paradigm for INRs. 
    Top: the underlying continuous signal $g$ and its derivatives $D_{\mathbf{x}}g$. 
    Middle: the pixels of image $\mathbf{I}$, which can be considered as discrete but exact sampling of $g$ and its derivatives are obtained by numerical approximation at image space. 
    Bottom: the MLP denoted as $f_{\boldsymbol{\Theta}}$ used as a general approximator whose derivatives are analytical thanks to auto-differentiation tools.
    In the proposed training paradigm, we enforce the supervision on both values and derivatives.
    }
    \label{fig:overview}
\end{figure}

Considering a continuous target signal $g: \mathbb{R}^D\rightarrow \mathbb{R}^C$, which is sampled discretely at $\{\mathbf{x}_i\}_{i=1}^N$, the goal of INRs is to approximate $g$ with a neural network (typically an MLP) $f_{\boldsymbol{\Theta}}: \mathbb{R}^D\rightarrow\mathbb{R}^C$, parameterized by weights $\boldsymbol{\Theta}$. 
The inputs to the MLP are typically low dimensional coordinates $\mathbf{x}\in\mathbb{R}^D$, e.g., $D=2$ for image coordinates, and the corresponding outputs are signal values, e.g., RGB colors.
Training pairs of most recent INRs are denoted as $\{(\mathbf{x}_i, g(\mathbf{x}_i))\}_{i=1}^{N}$, where $N$ is the number of total training samples.

By measuring the distance between predictions of the MLP $\{f_{\boldsymbol{\Theta}}(\mathbf{x}_i)\}_{i=1}^{N}$ and ground truth signal values $\{g(\mathbf{x}_i)\}_{i=1}^{N}$ with certain metrics, we actually obtain an optimization objective for tuning $\boldsymbol{\Theta}$.
Benefiting from the natural differentiability of MLPs, we are able to minimize the error using gradient descent (GD) implemented by auto differentiation frameworks.

MLPs are known as a powerful approximator of functions and with their continuous property, we are able to interpolate the discrete signal up to an arbitrarily higher resolution. 
However, given the traditional training scheme, the resulting ability to interpolate at unseen coordinates is usually not satisfactory. In this paper, we alleviate this problem by leveraging the supervision on derivatives, which is to say, we optimize $\boldsymbol{\Theta}$ by supervising not only signal values, but also derivatives at given coordinates. 

The overall training paradigm is illustrated in \cref{fig:overview}. We therefore construct a training dataset $\{(\mathbf{x}_i, g(\mathbf{x}_i), D_{\mathbf{x}}g(\mathbf{x}_i))\}_{i=1}^{N}$ for INRs. 
Since the analytical expression of $g$ is unknown, its derivatives require to be approximated with finite differences, which will be elaborated in \cref{sec:finite}. 
Considering that the task is a standard regression task, we apply $\ell$-2 error as the loss function.
With the training set, the optimization objective evolves to a combination of both supervision on signal values and first-order derivatives:
\begin{equation}\label{eq:general}
\begin{aligned}
      \boldsymbol{\Theta} & =  \mathop{\arg\min}\limits_{\boldsymbol{\Theta}}\frac1N\sum_{i=1}^{N}\left[
    \mathcal{L}_{\textrm{val}}(g(\mathbf{x}_i), f_{\boldsymbol{\Theta}}(\mathbf{x}_i))+\lambda \mathcal{L}_{\textrm{der}}(D_{\mathbf{x}}g(\mathbf{x}_i), D_{\mathbf{x}}f_{\boldsymbol{\Theta}}(\mathbf{x}_i))
    \right]\\
    & =  \mathop{\arg\min}\limits_{\boldsymbol{\Theta}}\frac1N\sum_{i=1}^{N}(
    \|g(\mathbf{x}_i)- f_{\boldsymbol{\Theta}}(\mathbf{x}_i)\|_2^2+\lambda \|D_{\mathbf{x}}g(\mathbf{x}_i)- D_{\mathbf{x}}f_{\boldsymbol{\Theta}}(\mathbf{x}_i)\|_2^2
    ).  
\end{aligned}
\end{equation}

We additionally extend this paradigm to enable both pre-processing of $\mathbf{x}$ and post-processing of $f_{\boldsymbol{\Theta}}(\mathbf{x})$, for instances that INRs are used to be an intermediate representation where the inputs to the MLP are calculated from $\mathbf{x}$ and the outputs of the MLP require further processing to obtain the final results. To formulate, we parameterize the MLP as $f_{\boldsymbol{\Theta}}$, and denote the pre-processing and post-processing as $p$ and $q$ respectively. The target signal $g$ is therefore approximated as $q(f_{\boldsymbol{\Theta}}(p(\mathbf{x})))$. Assuming that both $p$ and $q$ are differentiable and non-parametric, we can integrate these two processes into $f_{\boldsymbol{\Theta}}$ and follow the aforementioned training paradigm. We will carry out experiments on tasks satisfying this configuration in \cref{sec:indirect}. Specially, if $p$ and $q$ are identical mappings, the case degenerates to the original form.

Though can be applied under more circumstances, in this paper, we set the scope of target signals to $g: \mathbb{R}^2\rightarrow \mathbb{R}^3$, that $\{\mathbf{x}_i\}_{i=1}^N$ are image coordinates and $\{g(\mathbf{x}_i)\}_{i=1}^N$ are pixel values (colors). 
The loss supervision on derivatives $\mathcal{L}_{\textrm{der}}$ in \cref{eq:general} is composed of partial derivatives w.r.t. $u$ and $v$ respectively.

\subsection{Approximate Derivatives with Finite Differences}\label{sec:finite}
To enable Sobolev training on image coordinates, the first-order derivative of the target signal $g$ is supposed to be known as a precondition. However, the only information accessible at this stage is discrete sampled values $\{g(\mathbf{x}_i)\}_{i=1}^N$ so we need to approximate derivatives from these values with finite differences, or namely numerical differentiation (\cref{fig:overview}). Since we only consider cases that $\mathbf{x}\in \mathbb{R}^2$, the partial derivatives are 
\begin{equation}\label{eq}
\begin{aligned}
    D_{u}g(u,v) = & \frac{g(u+h,v)-g(u-h,v)}{2h},\\
    D_{v}g(u,v) = & \frac{g(u,v+h)-g(u,v-h)}{2h}.
\end{aligned}
\end{equation}

In this paper, we mainly apply the Sobel operator to obtain first-order derivatives of images. It defines a template that convolves the image $\mathbf{I}$ to obtain the partial derivative w.r.t. $u$ while the transposed template for the partial derivative w.r.t. $v$:
\begin{equation}
    D_u = \begin{bmatrix*}[r]
        -1 & \hphantom{+}0 & \hphantom{+}1\\
        -2 & 0 & 2\\
        -1 & 0 & 1
    \end{bmatrix*} * \mathbf{I},D_v = \begin{bmatrix*}[r]
        -1 & -2 & -1\\
        0 & 0 & 0\\
        1 & 2 & 1
    \end{bmatrix*} * \mathbf{I},
\end{equation}
where $*$ denotes the 2D convolution operation. We also study the choice of image filters for derivatives in \cref{sec:filter}, the conclusion of which is that enforcing the supervision of derivatives matters, regardless of the specific choice of filters.

\subsection{Activation Functions}

\begin{figure}[htbp]
    \centering
    \begin{subfigure}[t]{0.49\linewidth}
    \centering
    \includegraphics[width=\linewidth]{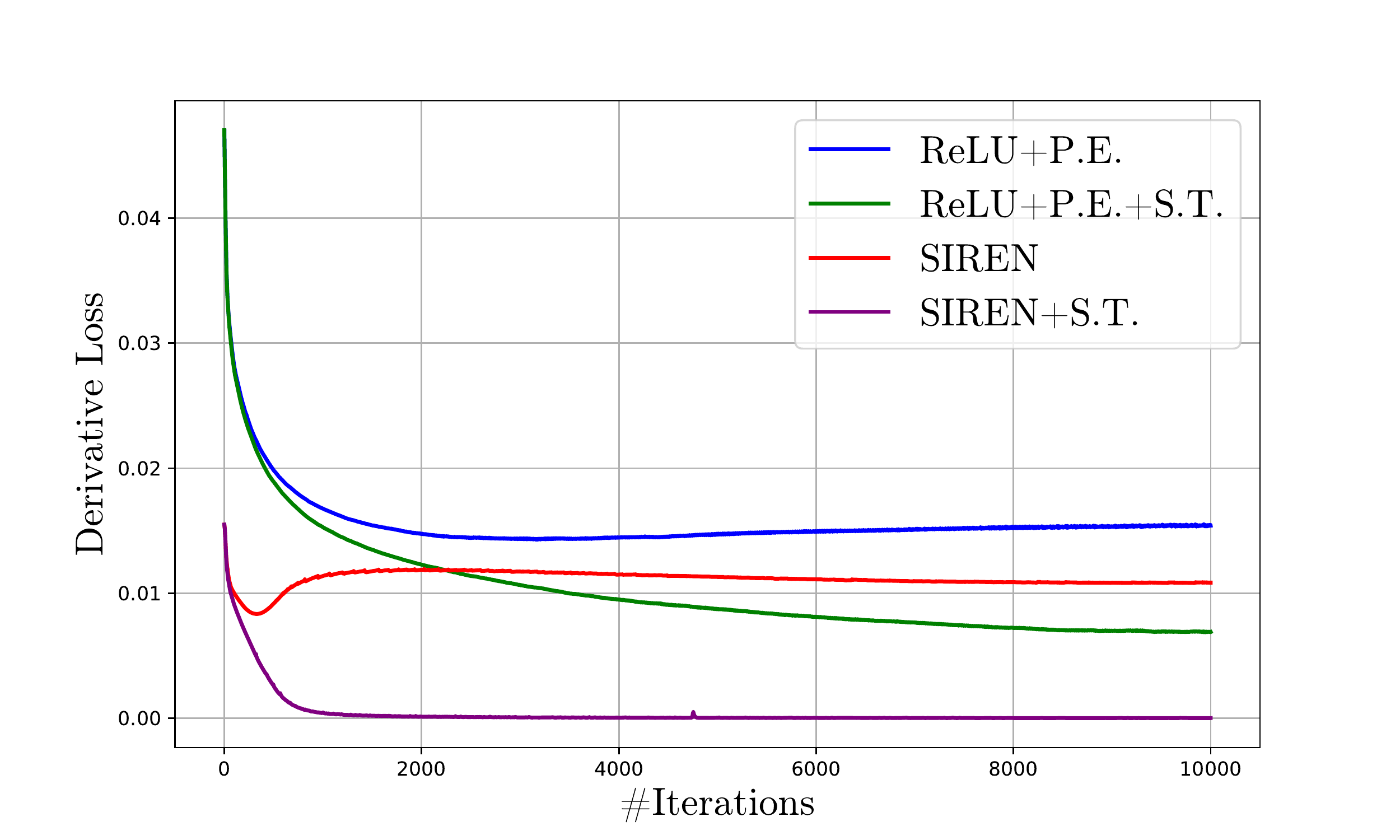}
    \caption{}
    \label{fig:act}
    \end{subfigure}
    \begin{subfigure}[t]{0.49\linewidth}
    \centering
    \includegraphics[width=\linewidth]{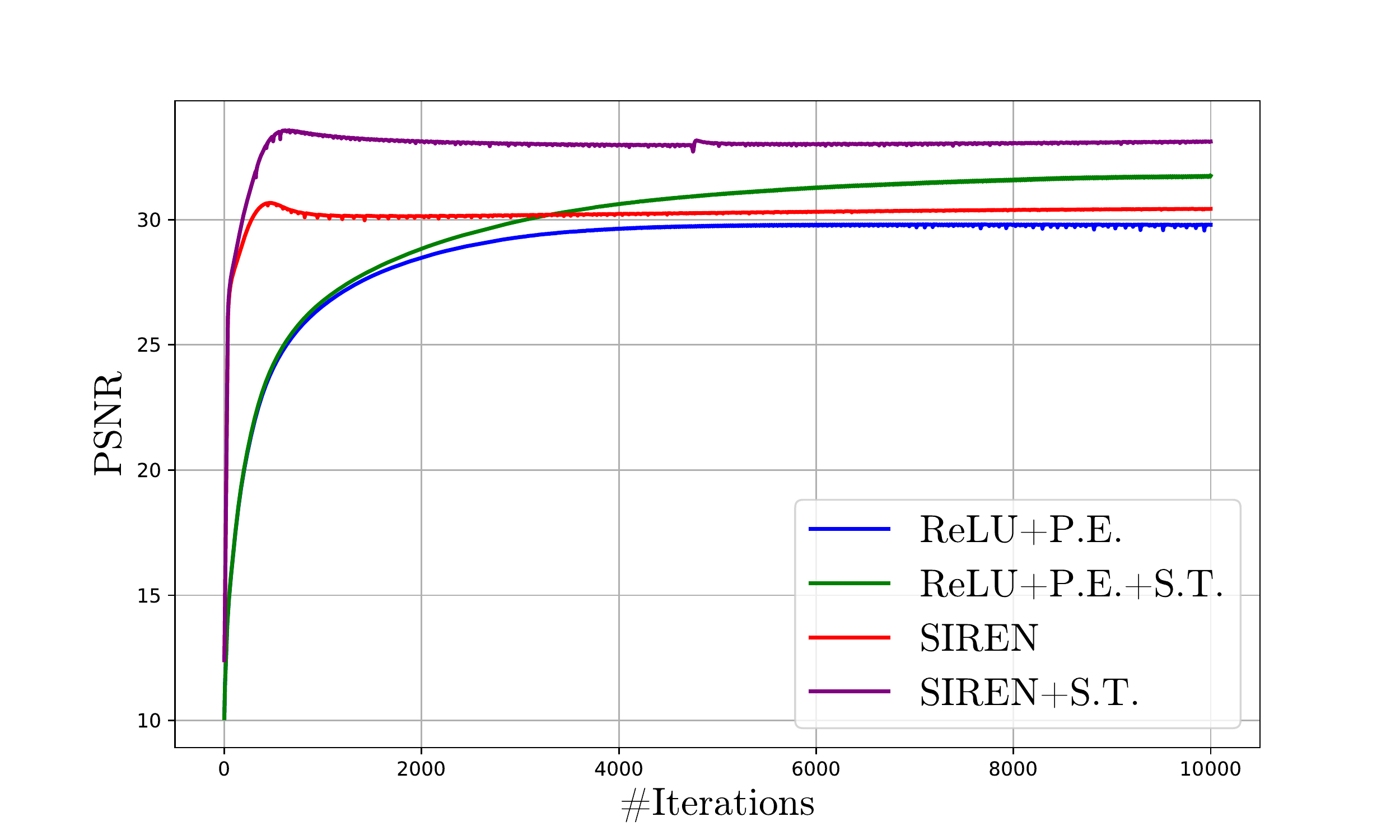}
    \caption{}
    \label{fig:curves}
    \end{subfigure}
    \caption{Convergence situations of $\mathcal{L}_{\textrm{der}}$ (a) and PSNR (b) w.r.t. iterations.}
\end{figure}

Activation functions are usually non-linear functions applied after each fully-connected layer. 
ReLU (Rectified Linear Units)~\cite{agarap2018deep} is a universally applied choice for its simple design, strong biological motivations and mathematical justifications. 
Due to the fact that the ReLU function is piecewise linear and its second derivative is zero everywhere, ReLU can not fit derivatives well in practice even the positional encoding is applied, just as shown in \cref{fig:act}.  
On the other hand, \cite{Parascandolo2017TamingTW,sitzmann2020implicit} attempt to explore the potential of the sine function as activation functions due to its periodicity, boundedness, arbitrary-order differentiability. 
The superior convergence property of sine has been proved in \cite{sitzmann2020implicit} when supervising the derivative only. 
In our setting, we try to replace ReLU with sine for better convergence property in Sobolev space. As can be indicated from \cref{fig:act}, sine indeed converges to a smaller derivative difference (close to 0) than ReLU when the training in Sobolev space is enabled.

However, studying the properties of activation functions in-depth is a complicated problem and in this paper, we mainly study the training paradigm applied to MLPs with activation functions of (a) ReLU, whose first-order derivative is piece-wise smooth, and (b) sine, whose arbitrary-order derivative is itself (with phase shift). We also study other activation functions in the Supplementary Material.
\section{Experiments}

In this section, we study the effectiveness of the proposed training paradigm on specific tasks that satisfy the preconditions. 
We sort tasks involving INRs into two categories, namely direct and indirect ones. 
Direct tasks are those where the supervision labels are in the same space as the network outputs, where $p$ and $q$ defined in \cref{sec:formulation} are both identical mappings.
While in indirect tasks, the network outputs are processed afterwards so that the observations in the same space as the supervision labels are produced. 
We give an example of direct and indirect tasks respectively in \cref{sec:direct} and \cref{sec:indirect}.

\subsection{Direct: Image Regression}\label{sec:direct}
A typical direct task is image regression. Given an RGB image $\mathbf{I}$, an MLP is appointed to regress the mapping from image coordinates to RGB colors $g: \mathbb{R}^2\rightarrow \mathbb{R}^3$. This task is fundamental regarding demonstrating the representation ability of INRs.

\subsubsection{Implementation}
\paragraph{Dataset}
We construct a dataset based on the Set 5 dataset~\cite{bevilacqua2012low} to demonstrate the great interpolatability of MLPs when trained with additional supervision on derivatives.
We split the pixels within an image into two groups, namely training samples and evaluation samples, by performing nearest-neighbor downscaling by a factor of 4. This is to say, for an image patch of $4\times 4$, we have 1 training sample and the remaining 15 pixels as evaluation samples.
\paragraph{Network Architecture}
For network architecture, we implement a fully-connected MLP with 4 activated linear layers with 256 hidden units each and 1 output linear layer. For the ReLU-based implementation, we follow the conclusion in \cite{yuce2021structured,tancik2020fourier} and apply positional encoding (P.E.) to the inputs (additional 20 channels).
\paragraph{Training \& Evaluation}
We train the network with all training samples with Adam optimizer for 50k iterations, at a learning rate of $10^{-4}$.
Following the common practice in image super resolution~\cite{jo2021practical,soh2019natural,yang2019deep}, we measure PSNR and SSIM on the Y channel and ignore 4 pixels from the image border. 
For PSNR, we only take the evaluation samples into account.

\begin{table}[t]
\setlength\tabcolsep{4pt}
\centering
\caption{Quantitative results on the task of image regression on Set 5 dataset~\cite{bevilacqua2012low}. When trained with additional supervision on derivatives and activated by sine and ReLU functions, the network achieves the best and second-best performance respectively in terms of both PSNR and SSIM. We also provided interpolated results by bilinear and bicubic interpolation. \bestcolor{best} \secondbestcolor{second-best}}
\label{tab:imgreg}
\begin{tabular}{c|c|ccccc}
\hline\hline
\textbf{Method} & \textbf{Mean} & \textit{Baby} & \textit{Bird} & \textit{Butterfly} & \textit{Head} & \textit{Woman}\\
\hline
\multicolumn{7}{c}{PSNR $\uparrow$}\\
\hline
Bilinear & 24.14 & 26.98 & 24.88 & 18.68 & 28.04 & 22.10\\
\hline
Bicubic & 23.63 & 26.52 & 24.46 & 18.23 & 27.43 & 21.52\\
\hline
ReLU+P.E.~\cite{mildenhall2020nerf} & 26.59 & 29.80 & 28.18 & 20.54 &29.47 & 24.95  \\ 
\hline
ReLU+P.E.+S.T. & \secondbestcolor{28.63} & \secondbestcolor{31.93} & \secondbestcolor{31.22} & \secondbestcolor{21.93} & \secondbestcolor{31.11} & \secondbestcolor{26.97} \\
\hline
SIREN~\cite{sitzmann2020implicit} & 26.22 & 30.48 & 28.07 & 20.73 & 28.22 & 23.62 \\
\hline
SIREN+S.T. & \bestcolor{30.28} & \bestcolor{33.36} & \bestcolor{33.34} & \bestcolor{24.42} & \bestcolor{31.64} & \bestcolor{28.62}\\
\hline\hline
\multicolumn{7}{c}{SSIM $\uparrow$}\\
\hline
Bilinear & 0.716 & 0.778 & 0.754 & 0.636 & 0.674 & 0.736\\
\hline
Bicubic & 0.705 & 0.773 & 0.746 & 0.625 & 0.655 & 0.727\\
\hline
ReLU+P.E.~\cite{mildenhall2020nerf} & 0.740 & 0.792 & 0.808 & 0.636 & 0.690 & 0.775  \\ 
\hline
ReLU+P.E.+S.T. & \secondbestcolor{0.809} & \secondbestcolor{0.867} & \secondbestcolor{0.888} & \secondbestcolor{0.708} & \secondbestcolor{0.757} & \secondbestcolor{0.827}\\
\hline
SIREN~\cite{sitzmann2020implicit} & 0.722 & 0.849 & 0.804 & 0.698 & 0.635 & 0.625\\
\hline
SIREN+S.T. & \bestcolor{0.890} & \bestcolor{0.918} & \bestcolor{0.955} & \bestcolor{0.870} & \bestcolor{0.795} & \bestcolor{0.910}\\
\hline\hline
\end{tabular}
\end{table}

\begin{table}[htbp]
\centering
\caption{Quantitative results on the task of inverse rendering on LLFF dataset~\cite{mildenhall2019local}. When trained with additional supervision on derivatives and activated by sine functions, the network achieves the best performance in terms of both PSNR and SSIM. \bestcolor{best} \secondbestcolor{second-best}} 
\label{tab:invrender}
\resizebox{\linewidth}{!}{
\begin{tabular}{c|c|cccccccc}
\hline\hline
\textbf{Method} & \textbf{Mean} & \textit{Fern} & \textit{Flower} & \textit{Fortress} & \textit{Horns} & \textit{Leaves} & \textit{Orchids} & \textit{Room} & \textit{T-Rex}\\
\hline
\multicolumn{10}{c}{PSNR $\uparrow$}\\
\hline
ReLU+P.E.~\cite{mildenhall2020nerf} & 23.42 &  21.82 & 25.03 & 27.89 & \secondbestcolor{24.75} & 18.14 & 18.92 & \secondbestcolor{27.22} & 23.62\\ 
\hline
ReLU+P.E.+S.T. & \secondbestcolor{23.74} & \secondbestcolor{22.99} & 25.27 & \secondbestcolor{27.90} &24.50 & \secondbestcolor{19.30} & \secondbestcolor{19.40} & 27.10 &23.49 \\
\hline
SIREN\cite{sitzmann2020implicit}+P.E. & 23.38 & 21.61 & \secondbestcolor{25.30} & 27.60 & 24.73 & 18.22 &18.99 & 26.89 & \secondbestcolor{23.69}\\
\hline
SIREN+P.E.+S.T. & \bestcolor{24.38} & \bestcolor{23.63} & \bestcolor{25.99} & \bestcolor{28.30} & \bestcolor{25.24} & \bestcolor{20.02} & \bestcolor{19.74} & \bestcolor{27.76} & \bestcolor{24.40}\\
\hline\hline
\multicolumn{10}{c}{SSIM $\uparrow$}\\
\hline
ReLU+P.E.~\cite{mildenhall2020nerf} & 0.706 & 0.651 & 0.755 & \secondbestcolor{0.769} & \secondbestcolor{0.729} & 0.535 & 0.544 & 0.877 & \secondbestcolor{0.791}\\ 
\hline
ReLU+P.E.+S.T. & \secondbestcolor{0.708} & \secondbestcolor{0.680} & \secondbestcolor{0.757} & 0.762 & 0.711 & \secondbestcolor{0.559} & \secondbestcolor{0.554} & \secondbestcolor{0.878} & 0.768 \\
\hline
SIREN~\cite{sitzmann2020implicit}+P.E. & 0.682 & 0.596 & 0.739 & 0.763 & 0.720 & 0.495 & 0.525&0.851 & 0.768\\
\hline
SIREN+P.E.+S.T. & \bestcolor{0.751} & \bestcolor{0.721} & \bestcolor{0.800} & \bestcolor{0.811} & \bestcolor{0.756} & \bestcolor{0.629} & \bestcolor{0.592} & \bestcolor{0.893} & \bestcolor{0.807}\\
\hline
\hline
\end{tabular}}
\end{table}

\subsubsection{Results}
\begin{figure}[htbp]
    \centering
    \includegraphics[width=\linewidth]{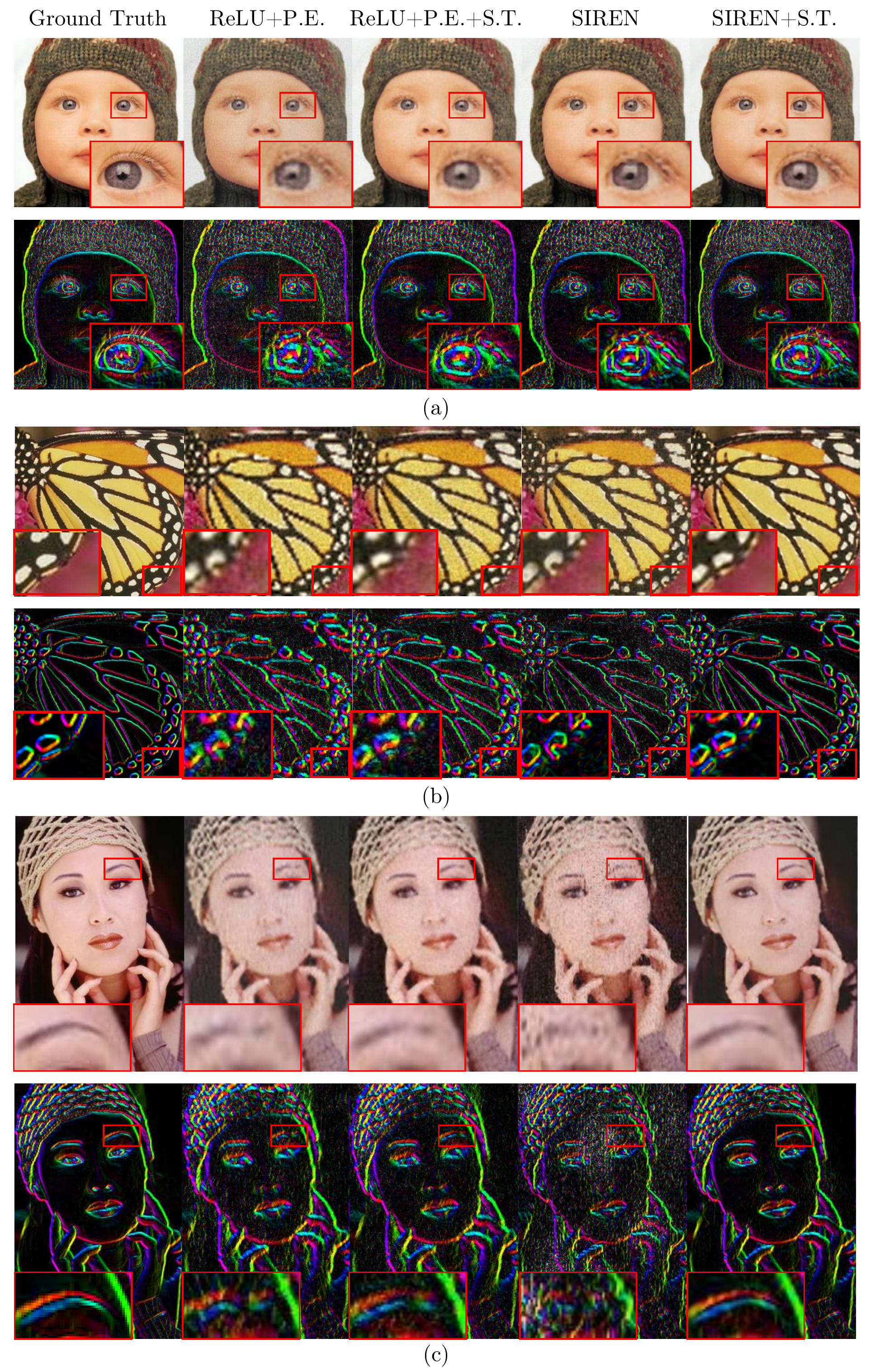}
    \caption{Results of regressed images and derivatives of the set \textit{Baby} (a), \textit{Butterfly} (b) and \textit{Woman} (c). When sine functions are adopted for activating linear layers and Sobolev training is enabled, the MLP yields the best visual effects, especially at high-frequency sharp edges, while ReLU-based MLPs fail to regress these details.}
    \label{fig:imgreg}
\end{figure}

The quantitative scores are shown in \cref{tab:imgreg}, where we compare the proposed training paradigm with the previous value-based training paradigm, and with different activation functions. When trained with $\mathcal{L}_{\textrm{val}}$ alone, ReLU and sine (SIREN~\cite{sitzmann2020implicit}) functions lead to comparable results; while with additional supervision on image derivatives, a huge performance improvement has been gained with both activations in both metrics.
It is worth noting that in all experiments, INRs demonstrate greater interpolatability than rule-based interpolation methods, i.e., bilinear and bicubic interpolation. 
\cref{fig:imgreg} gives some example regions where our method produces clearer images and sharper edges and also approximates better the derivatives. 
It can be observed that sine-based MLPs outperform ReLU-based ones when Sobolev training is enabled. Even so, Sobolev training still improves the results of ReLU-based INRs by a large margin. 
As additional evidences, \cref{fig:curves} shows the growth of PSNR w.r.t. the number of iterations, where the combination of Sobolev training and sine leads to the fastest convergence and the best performance.

\subsection{Indirect: Inverse Rendering}\label{sec:indirect}
INRs are recently attention-drawing for their great power demonstrated when representing 3D scenes. 
NeRF~\cite{mildenhall2020nerf}, which is a typical example of the kind, leverages an MLP $f_{\boldsymbol{\Theta}}: \mathbb{R}^5 \rightarrow \mathbb{R}^4$, to regress the mapping from $(x,y,z,\theta,\phi)$ to $(r,g,b,\sigma)$, where $x$, $y$, $z$ are 3D coordinates and $\theta$, $\phi$ are the view directions while the outputs are emitted colors $\mathbf{c}=(r,g,b)$ and volume density $\sigma$. 
In the pipeline of NeRF, the 3D sampling points $(x,y,z)$, as well as view directions $(\theta,\phi)$, of the MLP are determined from 2D image coordinates and corresponding calibrated camera parameters. 
After obtaining $(r,g,b,\sigma)$ for each sampling point, a numerical approximation of volume rendering is applied to obtain the integral colors. Considering that both pre-processing and post-processing are differentiable, we can adapt the extended formulation of proposed training paradigm mentioned in~\cref{sec:formulation} to perform inverse rendering of scenes.

\subsubsection{Implementation}
\paragraph{Dataset}
We carry out experiments on LLFF dataset~\cite{mildenhall2019local}. Similarly, we downsample the training images to $189\times 252$ while the resolution of target images to render is $756\times 1008$.
\paragraph{Network Architecture}
We implement a simplified version of NeRF~\cite{mildenhall2020nerf} $(x,y,z)\rightarrow (r,g,b,\sigma)$, removing the skip connection, the strategy of hierarchical sampling and view dependence. 
For network architecture, we implement a fully-connected MLP with 8 activated linear layers with 256 hidden units each and 1 output layer. 
We apply positional encoding (additional 60 channels) to both sine-activated and ReLU-activated MLPs.
\paragraph{Training \& Evaluation}
At the stage of training, we train the network with a batch size of 128 with Adam optimizer for 400k iterations, at a learning rate of $5\times 10^{-4}$. For evaluation metrics, we report PSNR and SSIM on all channels (RGB) of the rendered images.

\subsubsection{Results}

\begin{figure}[H]
    \centering
    \includegraphics[width=\linewidth]{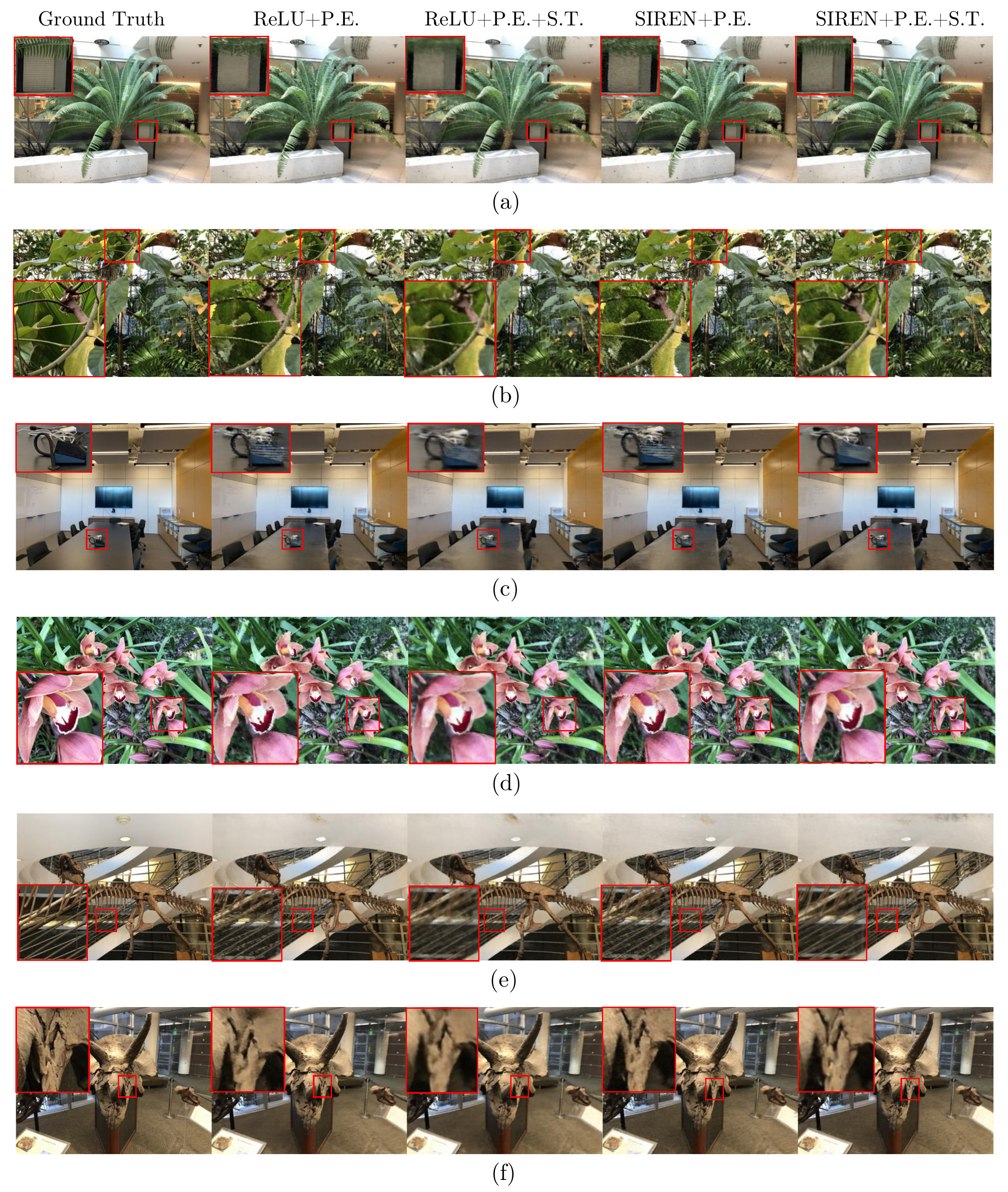}
    \caption{Results of novel view synthesis by INRs in the task of inverse rendering. Top to bottom: \textit{Fern} (a), \textit{Leaves} (b), \textit{Room} (c), \textit{Orchids} (d), \textit{T-Rex} (e), \textit{Horns} (f). Supervision on derivatives, applied with sine-activated MLPs, helps to render clear images and sharp edges at unseen views.}
    \label{fig:invrender}
\end{figure}

\cref{tab:invrender} demonstrates the PSNR and SSIM on all scenes of LLFF dataset~\cite{mildenhall2019local} and \cref{fig:invrender} provides some results of novel view synthesis. The conclusions drawn are similar to the task of image regression.

\paragraph{Data-efficiency} To provide more insights of the data-efficiency brought about by the proposed training paradigm, we also conduct experiments with different image resolution, namely different amount of training samples under different training paradigms. For the valued-based paradigm, we train MLPs with images of resolution $378\times 504$ and $756\times 1008$ respectively and render novel views of $756\times 1008$. 
For the proposed paradigm, we additionally train MLPs with images of $378\times504$ and render novel views of $756\times 1008$. 

The quantitative comparisons demonstrating the data-efficiency of Sobolev training are shown in \cref{tab:ratio}. 
The following two points can be observed and indicated. 
(a) Even with only $1/16$ training samples, the Sobolev trained MLP sometimes exceeds the ReLU-activated MLP with the value-based training paradigm, e.g., on \textit{Flower} in terms of PSNR and SSIM, on \textit{Fortress} and \textit{Horns} in terms of SSIM. 
(b) The Sobolev trained MLP using images of $378\times504$ gets similar PSNR and even higher SSIM compared to the ReLU-activated MLP trained with the value-based paradigm using images of $756\times 1008$. Note that the former case is trained only with $1/4$ samples as the latter.

\begin{table}[t]
\centering
\caption{Results for comparisons on the task of inverse rendering on LLFF dataset~\cite{mildenhall2019local} when the training resolution is larger. It is worth noting that even only trained with $1/16$ samples, the MLP trained with the proposed paradigm is able to outperform the ReLU-activated network on both metrics in the set of \textit{Flower}.}
\label{tab:ratio}
\resizebox{\linewidth}{!}{
\begin{tabular}{c|c|c|cccccccc}
\hline\hline
$H\times W$ & \textbf{Method} & \textbf{Mean} & \textit{Fern} & \textit{Flower} & \textit{Fortress} & \textit{Horns} & \textit{Leaves} & \textit{Orchids} & \textit{Room} & \textit{T-Rex}\\
\hline
\multicolumn{11}{c}{PSNR $\uparrow$}\\
\hline
$756\times 1008$ & ReLU+P.E. & \textbf{24.69} & 24.26 & 25.92 & \textbf{28.58} & 25.32 & 20.39 & \textbf{19.99} & \textbf{28.54} & 24.52\\
\hline
$378\times 504$ & ReLU+P.E. & 24.55 & 23.93 & 25.90 & 28.46 &25.22 &20.20 & 19.97 & 28.30 & 24.42\\
\hline
$378\times 504$ & SIREN+P.E.+S.T. & 24.67 & \textbf{24.27} & \textbf{26.16} & 28.37 & \textbf{25.37} & \textbf{20.64} & 19.94 & 27.99 & \textbf{24.61} \\
\hline
$189\times 252$ & SIREN+P.E.+S.T. & 24.38 & 23.63 & 25.99 & 28.30 & 25.24 & 20.02 &19.74 & 27.76& 24.40\\
\hline\hline
\multicolumn{11}{c}{SSIM $\uparrow$}\\
\hline
$756\times 1008$ & ReLU+P.E. & 0.755 & 0.744 & 0.788 & 0.792 & 0.746 & 0.646 & 0.606 & \textbf{0.900} & \textbf{0.816}\\
\hline
$378\times 504$ & ReLU+P.E. & 0.750 & 0.734 & 0.785 &0.790 & 0.744 & 0.637 & 0.603 & 0.897 & 0.813\\
\hline
$378\times 504$ & SIREN+P.E.+S.T. & \textbf{0.766} & \textbf{0.753} & \textbf{0.805} & \textbf{0.815} & \textbf{0.762} & \textbf{0.666} & \textbf{0.614} & 0.899 & 0.815 \\
\hline
$189\times 252$ & SIREN+P.E.+S.T. & 0.751 & 0.721 & 0.800 & 0.811 & 0.756 & 0.629 & 0.596 & 0.893 & 0.807\\
\hline
\hline
\end{tabular}}
\end{table}

\subsection{Ablation Study}
\subsubsection{Image Derivative Filters} \label{sec:filter}
We compare the network performance with another image filter as finite differences approximating derivatives, namely vanilla image derivative filter whose templates are
\begin{equation}
    T_u = \begin{bmatrix*}[r]
        0 & \hphantom{+}0 & \hphantom{+}0\\
        -1 & 0 & 1\\
        0 & 0 & 0
    \end{bmatrix*} ,T_v = \begin{bmatrix*}[r]
        \hphantom{+}0 & -1 & \hphantom{+}0\\
        0 & 0 & 0\\
        0 & 1 & 0
    \end{bmatrix*}.
\end{equation}
It considers the same neighborhood as Sobel operator but removes the spatial smoothing operation included in Sobel operator. The quantitative comparisons are shown in \cref{tab:filter}. 

Though the results with Sobel operator are slightly better than the vanilla image derivative filter, the performance gap is relatively small, indicating the specific choice of filters is not crucial to the performance. 

\begin{table}[t]
\setlength\tabcolsep{2pt}
\begin{minipage}[t]{0.48\linewidth}
\centering
\caption{Quantitative comparison on the choice of image filters when approximating derivatives of $g$. Experiments are carried out with MLPs activated by sine functions. PSNR and SSIM stand for mean values over all scenes.}
\label{tab:filter}
\begin{tabular}{c|c|cc}
\hline\hline
Task & Filter & PSNR & SSIM \\
\hline
Image & Vanilla & 30.24 & 0.886\\
\cline{2-4}
Regression & Sobel & \textbf{30.28} &  \textbf{0.890} \\
\hline
Inverse & Vanilla & 24.28 & 0.746\\
\cline{2-4}
Rendering & Sobel & \textbf{24.38}& \textbf{0.751}\\
\hline
\hline
\end{tabular}
\end{minipage}\hfill
\begin{minipage}[t]{0.48\linewidth}
    \centering
    \caption{Quantitative comparison on the number of layers in MLPs under the task of inverse rendering. PSNR and SSIM stand for mean values over all scenes.}
    \label{tab:layer}
    \resizebox{\linewidth}{!}{
    \begin{tabular}{c|c|cc}
\hline\hline
\#layers & \textbf{Method} & PSNR & SSIM\\
\hline
\multirow{4}*{4} & ReLU+P.E. & 23.29 &  0.685 \\ 
\cline{2-4}
 & SIREN+P.E. & 23.17 & 0.654 \\
\cline{2-4}
& ReLU+P.E.+S.T. & 23.35 & 0.689\\
\cline{2-4}
& SIREN+P.E.+S.T. & 23.66 & 0.697\\
\hline
\multirow{4}*{8} & ReLU+P.E. & 23.42 & 0.706\\ 
\cline{2-4}
 & SIREN+P.E. & 23.38 & 0.682\\
\cline{2-4}
& ReLU+P.E.+S.T. & 23.74 & 0.708\\
\cline{2-4}
& SIREN+P.E.+S.T. & \textbf{24.38} & \textbf{0.751} \\
\hline\hline
\end{tabular}}
\end{minipage}
\end{table}

\subsubsection{Network Capacity}
A simple intuition towards the fitting ability, including both values and derivatives, of a neural network is that the more parameters a network possesses, the more complex functions it can represent.
We present the quantitative results obtained with networks with different number of layers (4 and 8 respectively) in the task of inverse rendering in \cref{tab:layer}. Other experiment settings are aligned with \cref{sec:indirect}.
It can be indicated that when the layer number increases to 8 from 4, the representation ability of INRs is stronger so it can obtain more performance gain from derivatives.

\section{Discussions}

\subsection{Future Work}

To restate, the preconditions of the proposed training paradigm are that (a) the raw dataset is in the form of $\{(\mathbf{x}_i,g(\mathbf{x}_i))\}_{i=1}^N$, where $\{\mathbf{x}_i\}_{i=1}^N$ are spatial coordinates; (b) the derivatives of $g$ can be obtained or approximated with limited $\{g(\mathbf{x}_i)\}$; (c) $\{\mathbf{x}_i\}$ do not necessarily be the direct inputs of the MLP, and the target values $\{g(\mathbf{x}_i)\}$ do not necessarily be the direct outputs if and only if the pre-processing $p$ is differentiable w.r.t. to $\mathbf{x}$ and the post-processing $q$ is differentiable w.r.t. $f_{\boldsymbol{\Theta}}(p(\mathbf{x}))$.

Theoretically, referring to the formulation in \cref{sec:formulation}, the training paradigm can further generalize to more tasks, e.g., audio regression (please refer to the Supplementary Material) and video regression, as long as their partial derivatives can be obtained by numerical approximation. 
Take the task of video regression $(u,v,t)\rightarrow (r,g,b)$ as an example, whose partial derivatives w.r.t. $u$ and $v$ are the same on images. We can approximate the partial derivative w.r.t. $t$ by subtracting consecutive frames.

\subsection{Limitations}
We summarize the limitations of the proposed training paradigm as below.
\begin{enumerate}
\setlength{\itemsep}{0pt}
\setlength{\parsep}{0pt}
\setlength{\parskip}{0pt}
    \item[-] The computation of analytical derivatives is computationally expensive and occupies a considerable amount of memory. For example, both PyTorch~\cite{paszke2019pytorch} and TensorFlow~\cite{abadi2016tensorflow} cannot calculate derivatives for each sample in a batch separately, leading to redundant computation. 
    \item[-] Though the proposed training paradigm can be applied to various INRs-based tasks, the scope of application is still limited. Since the derivatives are approximated with finite differences, the distribution of $\{\mathbf{x}_i\}$ is supposed to be as uniform as possible at its domain, which is not always satisfied.
\end{enumerate}

\section{Conclusion}
In this paper, we put forward a novel training paradigm for INRs in Sobolev space, where supervision on both signal values and first-order derivatives is enforced for network training. 
The formulation of the task is generalized to any circumstance that the inputs are 2D image coordinates and the outputs are image values (e.g., RGB) as long as the pre-processing is differentiable w.r.t. the coordinates and the post-processing is differentiable w.r.t. the MLP's outputs. We obtain approximated image derivatives by the Sobel operator.

Experiments are carried out on the task of image regression (direct) and inverse rendering (indirect) and results show that the training paradigm brings remarkable improvement to the quality of reconstructed images, especially when applied together with an MLP whose activation functions are sine.

In addition, we study some important and interesting peripheral problems relevant to the proposed training paradigm and give insights based on our observations. 

\subsubsection{Acknowledgements}
The authors would like to thank Zhongtian Zheng from Peking University for his valuable advice.

%\clearpage
\bibliographystyle{splncs04}
\bibliography{egbib}

\appendix
\clearpage
\section{Additional Details \& Results}

\subsection{Image Regression}

\begin{figure}[t]
	\centering
	\includegraphics[width=0.9\linewidth]{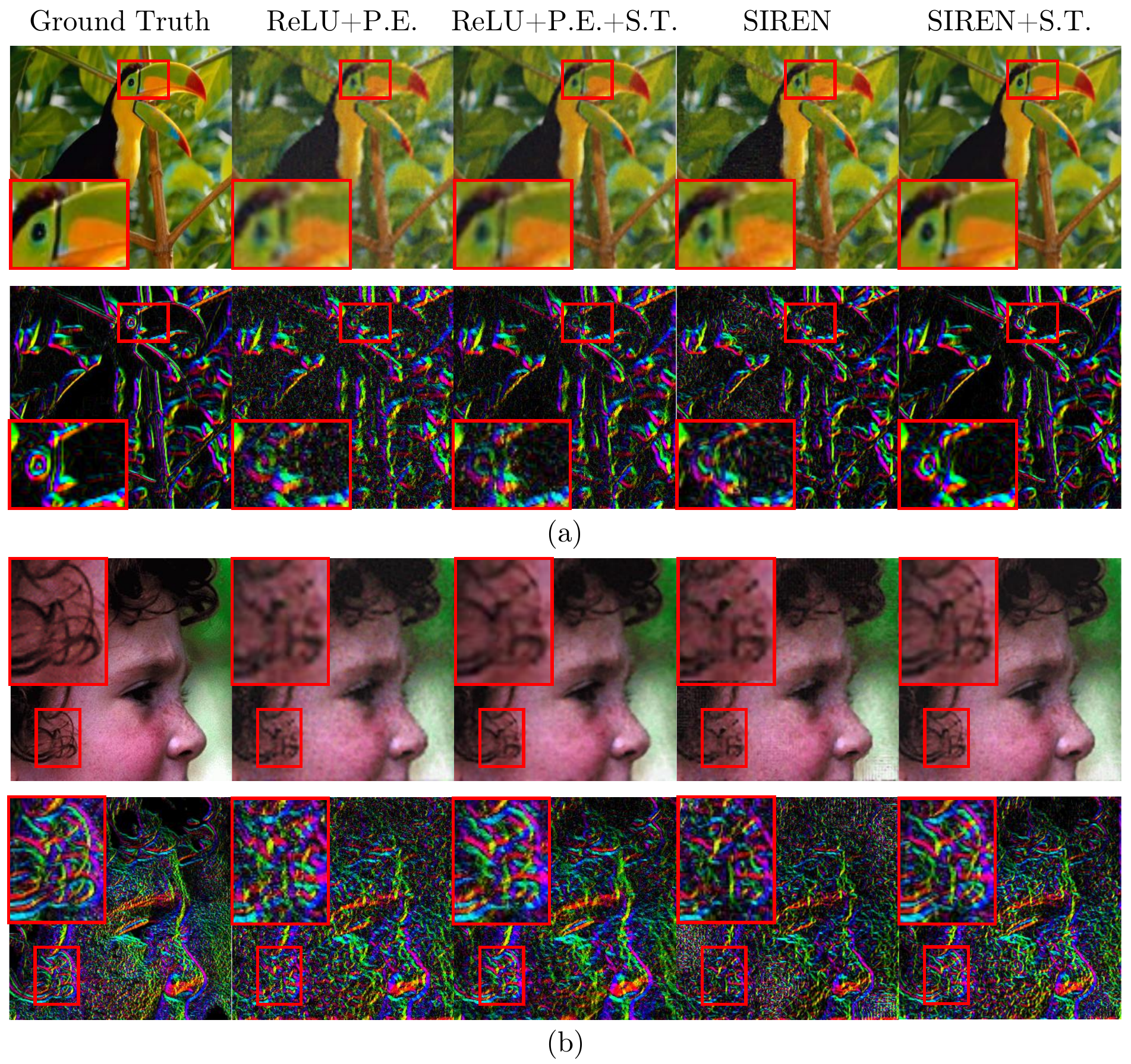}
	\caption{Additional results of image regression. \textit{Bird} (a), \textit{Head} (b).}
	\label{fig:compare_1}
\end{figure}

For image regression task, we set the angular velocity of the sinusoidal function to $30$ and initialize the weights of MLPs following \cite{sitzmann2020implicit}.
We provide other results of Set 5 dataset~\cite{bevilacqua2012low} in \cref{fig:compare_1}. We also report the memory consumption as well as training time required in \cref{tab:baby_memory_time}. The proposed training paradigm requires more memory and time at the training phase for additional computation and storage of derivatives but it is worth noting that the time, as well as the memory consumption at the inference phase, is the same with or without the derivative supervision.

\subsection{Inverse Rendering}
For inverse rendering task, we set the angular velocity of the sinusoidal function to $1$ and initialize the weights of MLPs according to the method described in~\cite{he2015delving}.
LLFF dataset~\cite{mildenhall2019local,mildenhall2020nerf} consists of 8 scenes captured with a handheld cellphone, captured with 20 to 62 images. We follow~\cite{mildenhall2020nerf} to hold out $\frac{1}{8}$ of images for the evaluation set. All of the training images are $756\times1008$, but the dataset also provides the raw cellphone images of $3024\times4032$, which will be used to evaluate high resolution rendering results in \cref{sec:precision}.

\begin{figure}[t]
	\centering
	\includegraphics[width=\linewidth]{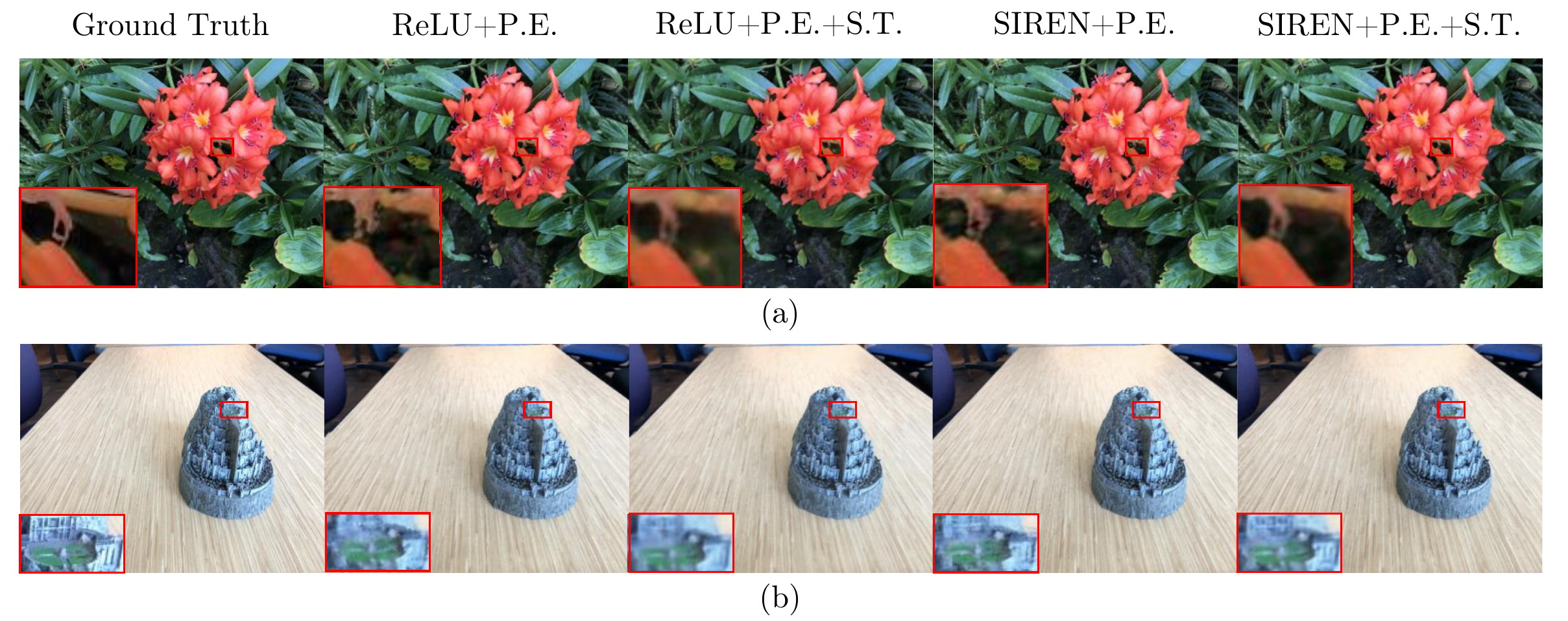}
	\caption{Additional results of inverse rendering. \textit{Flower} (a), \textit{Fortress} (b).}
	\label{fig:compare_2}
\end{figure}

We provide other results of LLFF dataset~\cite{mildenhall2019local,mildenhall2020nerf} in \cref{fig:compare_2}. \cref{tab:fern_memory_time} reports the training-time memory consumption and time. Similar to image regression, the additional cost is only at training-time.

\section{Audio Regression}
As mentioned in the discussions of the main paper, the training paradigm we proposed can further generalize to more tasks, as long as the tasks satisfy the formulation in Sec. 3.1 of the main paper. Here we follow \cite{sitzmann2020implicit} to conduct the task of audio signal representation.
\subsubsection{Implementation}
\paragraph{Dataset}
Following~\cite{sitzmann2020implicit}, we use two different audio signals, one for music and one for speech. For music data, we use the first 7 seconds from Bach's Cello Suite No.1 (\textit{Bach})\footnote{Audio file available at \url{https://www.yourclassical.org/episode/2017/04/04/daily-download-js-bach--cello-suite-no-1-prelude}.}, and for the speech, we use stock audio of a male actor counting from 0 to 9 (\textit{Counting})\footnote{Audio file available at \url{http://soundbible.com/2008-0-9-Male-Vocalized.html}.}.
Both audio signals share a sampling rate of 44100$Hz$, and in total there are 308207 samples in \textit{Bach} and 537936 samples in \textit{Counting}. We normalize signal values to the range of $[-1, 1]$.
Same as other tasks, we separate all samples within an audio clip into two groups, namely training samples and evaluation samples, by performing nearest-neighbor downscaling by a factor of 5. The approximated derivatives are obtained by two-sided differences method, similar to the vanilla derivative filter of the task on images.

\paragraph{Network Architecture}
We implement a fully-connected MLP with 4 activated linear layers with 256 hidden units and 1 output linear layer. We set the angular velocity of the sinusoidal function to $30$ and follow the initialization strategy in \cite{sitzmann2020implicit} to initialize the weights of MLPs.
\paragraph{Training \& Evaluation}
At the stage of training, we use the entire set of training samples to train the network as a batch with Adam optimizer, at a learning rate of $5\times 10^{-5}$. For evaluation, we measure PSNR only on the evaluation samples.
\subsubsection{Results}
\begin{figure}[t]
	\centering
	\begin{subfigure}[t]{\linewidth}
		\centering
		\includegraphics[width=0.8\linewidth]{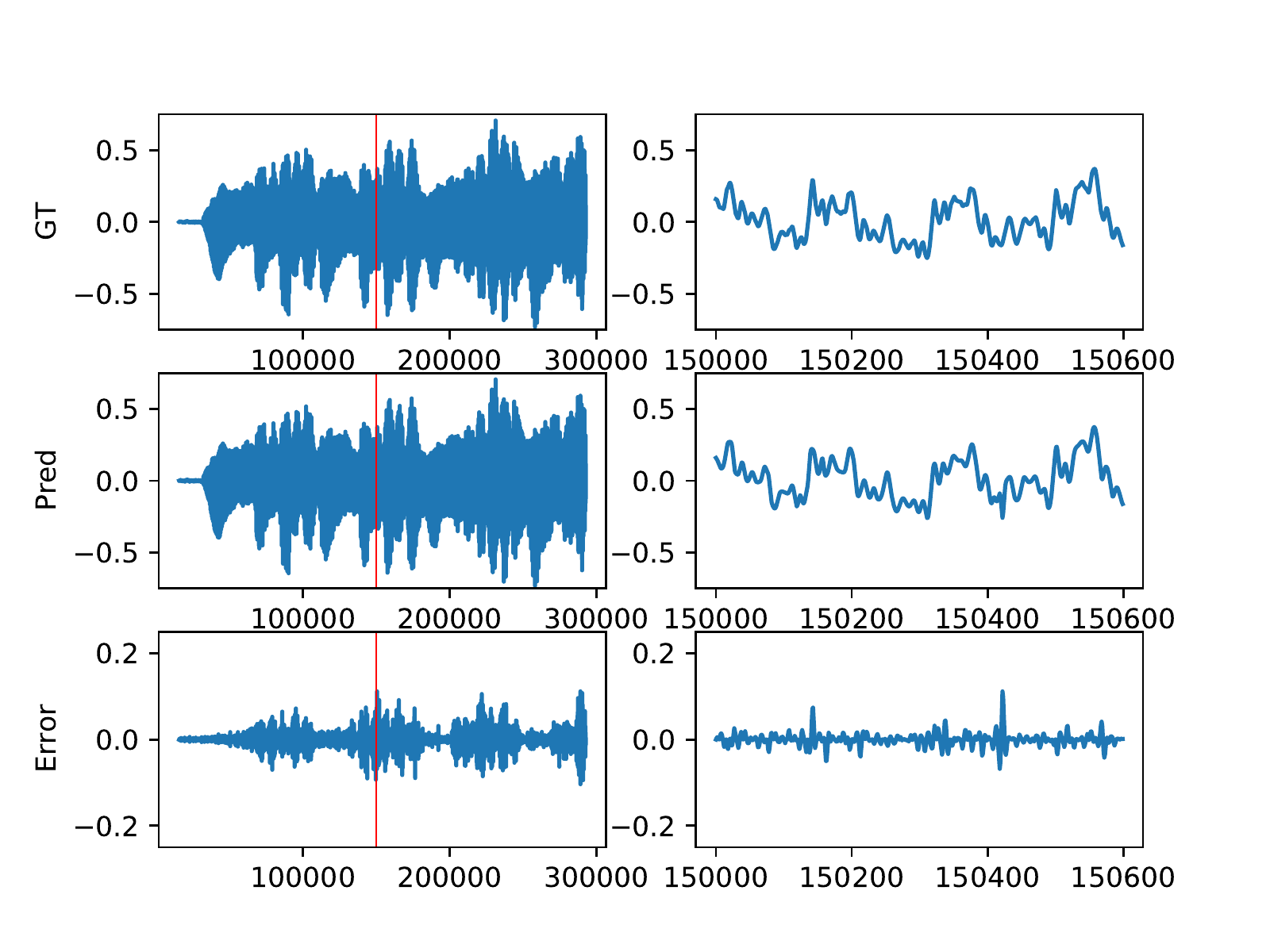}
		\caption{SIREN~\cite{sitzmann2020implicit}}
	\end{subfigure}
	\begin{subfigure}[t]{\linewidth}
		\centering
		\includegraphics[width=0.8\linewidth]{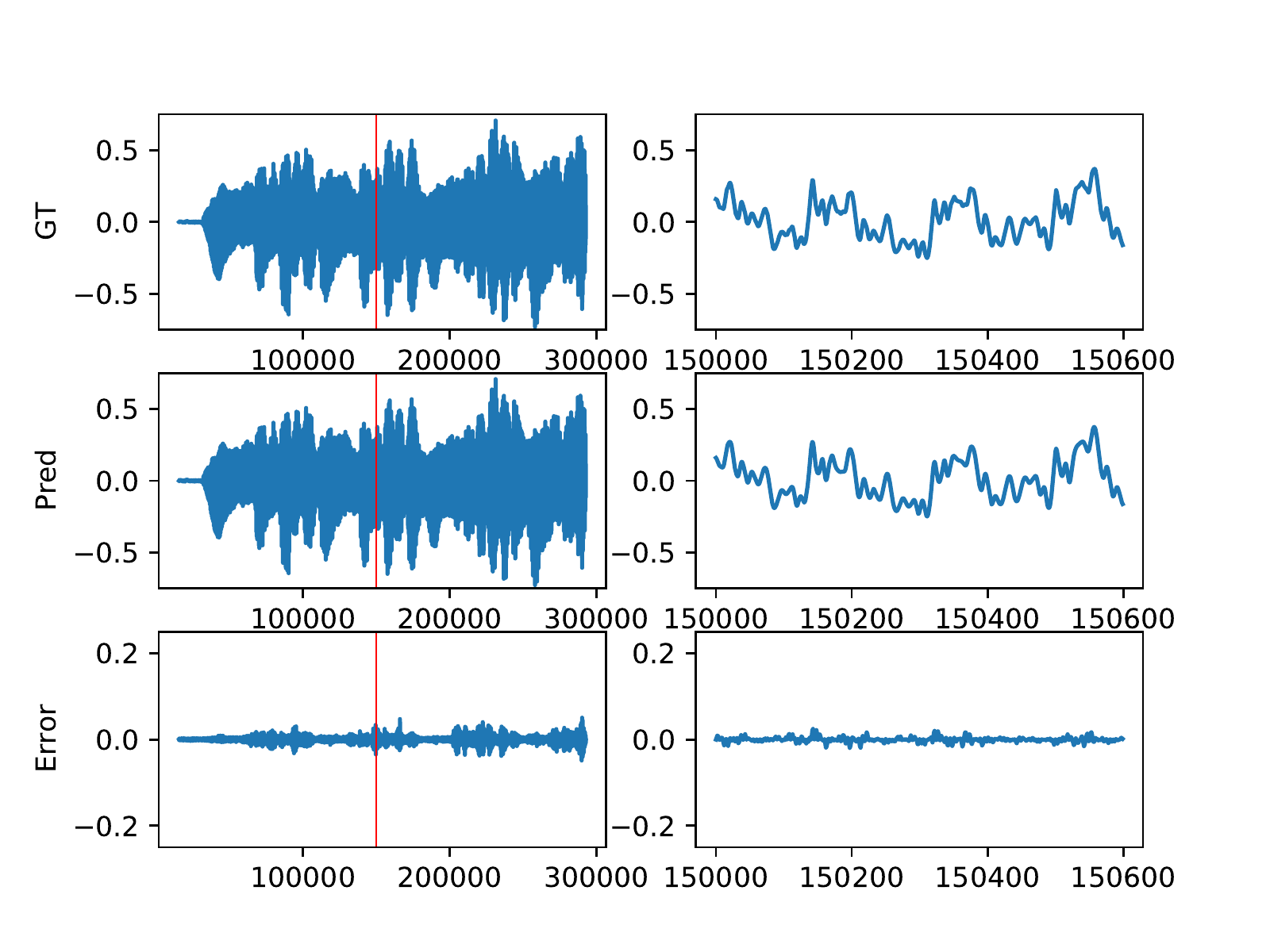}
		\caption{SIREN+S.T.}
	\end{subfigure}
	\caption{Regressed waveforms of the INRs trained with and without derivative supervision on the clip of \textit{Bach}. From top to bottom: ground truth waveform, regressed waveform and error detected. We zoom in the waveform marked with red at the right column accordingly.}
	\label{fig:bach}
\end{figure}

\begin{figure}[t]
	\centering
	\begin{subfigure}[t]{\linewidth}
		\centering
		\includegraphics[width=0.8\linewidth]{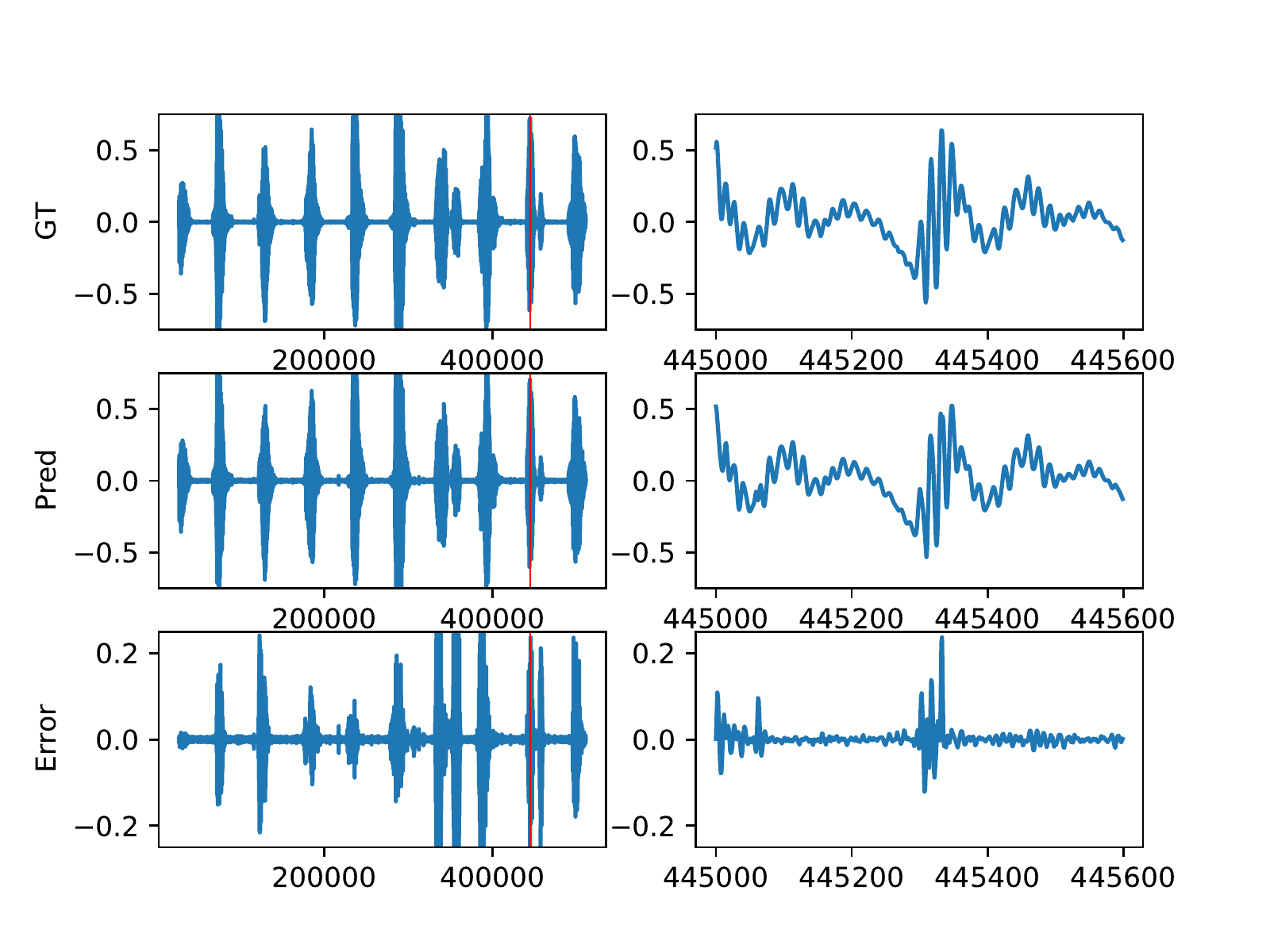}
		\caption{SIREN~\cite{sitzmann2020implicit}}
	\end{subfigure}
	\begin{subfigure}[t]{\linewidth}
		\centering
		\includegraphics[width=0.8\linewidth]{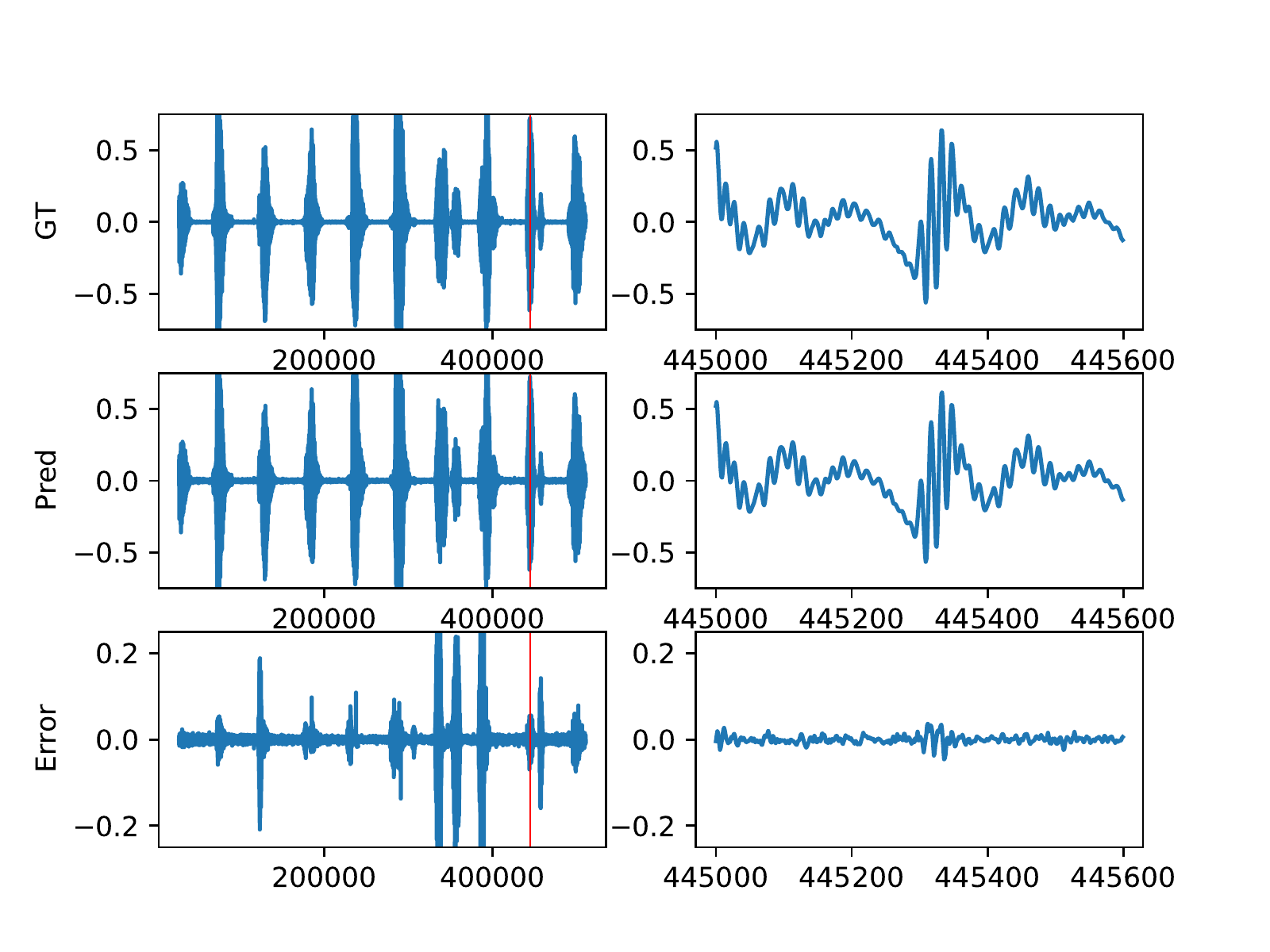}
		\caption{SIREN+S.T.}
	\end{subfigure}
	\caption{Regressed waveforms of the INRs trained with and without derivative supervision on the clip of \textit{Counting}. From top to bottom: ground truth waveform, regressed waveform and error detected. We zoom in the waveform marked with red at the right column accordingly.}
	\label{fig:counting}
\end{figure}

The quantitative results in PSNR of regressing these two audio clips are shown in \cref{tab:audioreg}. 
Sobolev training significantly improves the regression quality of both audio clips. We further illustrate the visualized comparisons respectively in \cref{fig:bach} and \cref{fig:counting}, showing the regression error of the derivative-trained network is smaller.

As our results show ReLU-activated networks can not fit audio signals well, which is consistent with the result of~\cite{sitzmann2020implicit}, we only report the results of Sine-activated networks.

\begin{table}[t]
	\setlength\tabcolsep{4pt}
	\centering
	\caption{GPU memory consumption at training phase and training time of 1000 epochs in image regression task of \textit{Baby}. P.E. stands for positional encoding; S.T. stands for Sobolev training.}
	\label{tab:baby_memory_time}
	\begin{threeparttable}{
			\begin{tabular}{c|c|c}
				\hline\hline
				\textbf{Method} & Memory(\textit{MB}) & Time(\textit{s}) \\
				
				\hline
				ReLU+P.E.~\cite{mildenhall2020nerf} & 130 & 4.678 \\
				\hline
				ReLU+P.E.+S.T. & 316 & 20.548 \\
				\hline
				ReLU+P.E.+S.T.$^*$ & 401 &  14.081 \\
				\hline
				SIREN~\cite{sitzmann2020implicit} & 176 & 3.977 \\
				\hline
				SIREN+S.T. & 753 &  22.963 \\
				\hline
				SIREN+S.T.$^*$ & 704 &  16.556 \\
				\hline
				\hline
			\end{tabular}
			\begin{tablenotes}
				\item[*] The new PyTorch 1.11.0 supports computation of per-sample derivatives, so we also report the statistics obtained.
			\end{tablenotes}
		}
	\end{threeparttable}
\end{table}

\begin{table}[htb]
	\setlength\tabcolsep{4pt}
	\centering
	\caption{GPU memory consumption at training phase and training time of 1000 iterations in inverse rendering task of \textit{Fern}.}
	\label{tab:fern_memory_time}
	\begin{tabular}{c|c|c}
		\hline\hline
		\textbf{Method} & Memory(\textit{MB}) & Time(\textit{s}) \\
		
		\hline
		ReLU+P.E. & 228 & 87.264 \\
		\hline
		ReLU+P.E.+S.T. & 912 & 190.434 \\
		\hline
		SIREN+P.E. & 348 & 86.642 \\
		\hline
		SIREN+P.E.+S.T. & 1827 & 204.748 \\
		\hline
		\hline
	\end{tabular}
\end{table}

\begin{table}[t]
	\setlength\tabcolsep{4pt}
	\centering
	\caption{Quantitative PSNR results on the task of audio regression on \textit{Bach} and \textit{Counting}. $1/5$ samples are used for training, and rest samples are used for evaluation. When trained with additional supervision on derivatives, PSNR is significantly improved.}
	\label{tab:audioreg}
	\begin{tabular}{c|c|cc}
		\hline\hline
		\textbf{Method} & \textbf{Mean} & \textit{Bach} & \textit{Counting}\\
		\hline
		SIREN~\cite{sitzmann2020implicit} & 35.89 & 41.50 & 30.28\\
		\hline
		SIREN+S.T. & \textbf{41.23} & \textbf{48.89} & \textbf{33.57} \\
		\hline
		\hline
	\end{tabular}
\end{table}

\section{Different Activation Functions}
Recent ReLU-based MLPs normally have poor convergence property under derivative supervision, we mainly investigate the performance difference between ReLU and periodic activation function, i.e., sine, in the main paper. 
In this part, an experiment is designed to study the convergence properties of derivatives with different activation functions. The scope of the investigation is shown in \cref{tab:activations}.

\paragraph{Dataset} We convert the RGB image \textit{Bird} of Set 5~\cite{bevilacqua2012low} to grayscale as our dataset. The original image shape is $288\times288\times1$, we use the Sobel operator to get approximate image partial derivatives, resulting the derivative image with shape $288\times288\times2$, w.r.t. $u$ and $v$ respectively.
We perform nearest-neighbor downsampling on the original image and derivative image by a factor of 4.
\paragraph{Network Architecture} The network architecture is the same as the image regression task in the main paper, except for the channel number of the last linear layer is 1 instead of 3. The angular velocity of the sinusoidal function and weight initialization method are both the same as image regression task.
Taking the conclusion in \cite{tancik2020fourier,yuce2021structured} into consideration, we do two parallel experiments with and without positional encoding for all activation functions except for sine.

\paragraph{Training \& Evaluation}  
We use the downsampled $1/16$ pixels as our training data and the remaining pixels as evaluation samples. The network is optimized for 10k iterations at a learning rate of $10^{-4}$.

\subsubsection{Results}
\begin{figure}[t]
	\centering
	\includegraphics[width=\linewidth]{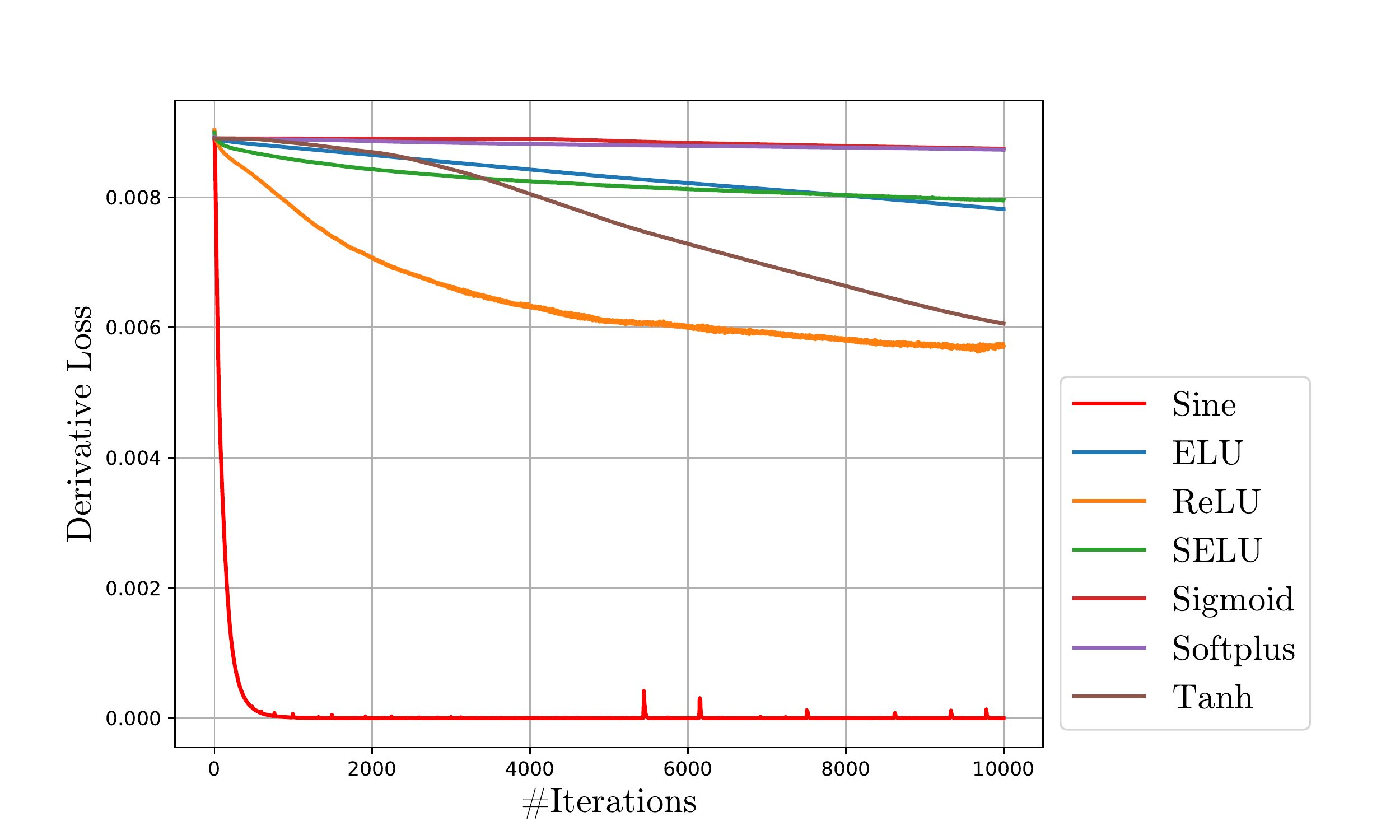}\vspace{-10pt}
	\caption{Derivative loss of different activation functions.}
	\label{fig:gradloss}
\end{figure}

\begin{figure}[t]
	\centering
	\includegraphics[width=\linewidth]{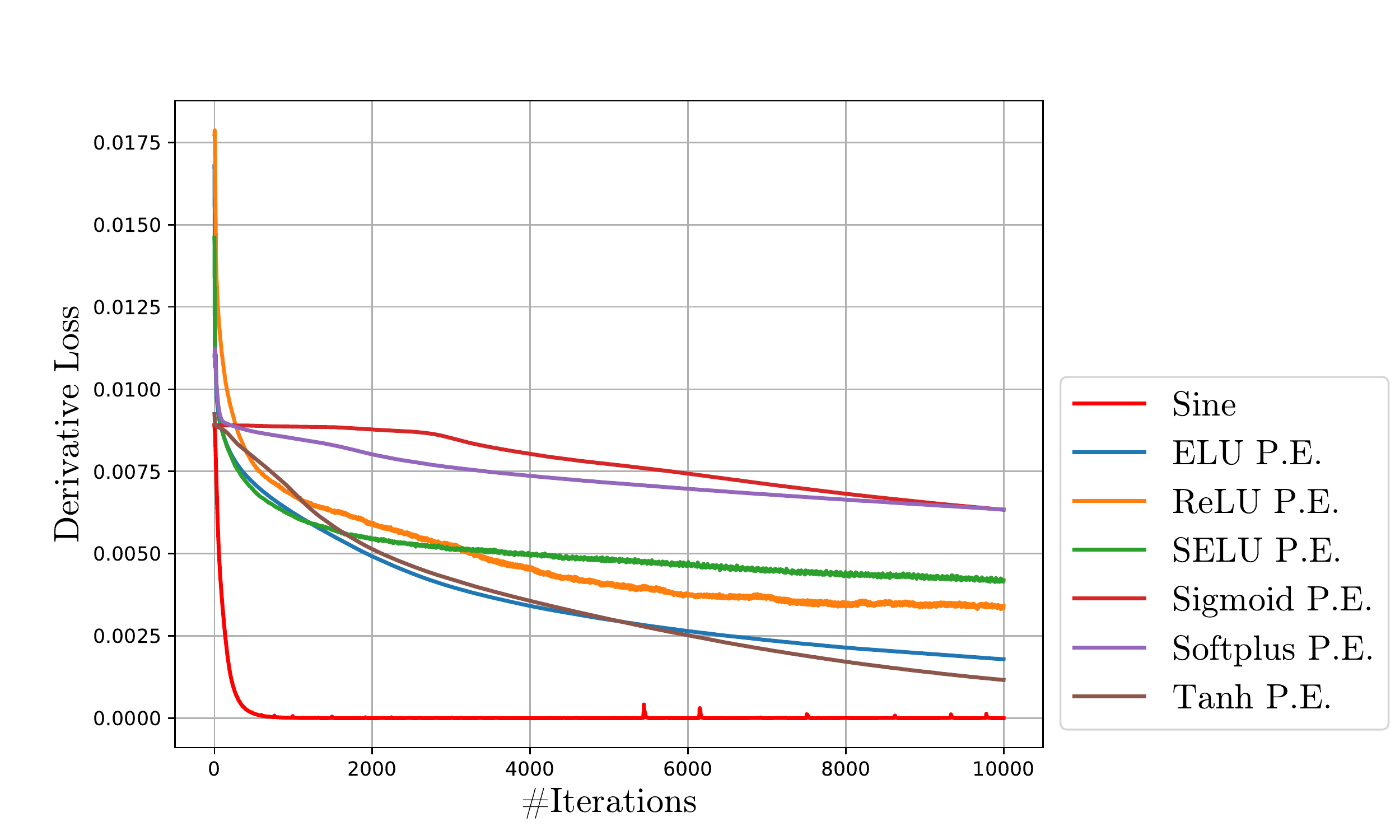}\vspace{-10pt}
	\caption{Derivative loss of different activation functions with positional encoding. P.E. means positional encoding.}
	\label{fig:gradlossPE}
\end{figure}

\cref{fig:gradloss} and \cref{fig:gradlossPE} respectively show the convergence curve of derivative loss with and without positional encoding. Sine demonstrates its great power in fitting the network's derivative compared with other activation functions. It is worth noting that positional encoding is helpful to improve the MLPs' ability of approximating derivatives when applied with other activation functions. The corresponding quantitative results are shown in~\cref{tab:grayreg}.

\begin{table}[t]
	\setlength\tabcolsep{4pt}
	\centering
	\caption{Different activation functions and their definitions and weight initialization}
	\label{tab:activations}
	\resizebox{\linewidth}{!}{
		\begin{tabular}{c|c|c|c}
			\hline\hline
			Activation & Definition & Derivative & Initialization\\
			
			\hline
			% Abs & $f(x)=\begin{cases}x,&x>0\\-x,&x\leq0\end{cases}$ & $f'(x)\begin{cases}1,&x>0\\-1,&x<0\end{cases}$ & Kaiming normal\cite{he2015delving}  \\
			% \hline
			ReLU~\cite{agarap2018deep} & $f(x)=\begin{cases}x,&x>0\\0,&x\leq0\end{cases}$ & $f'(x)=\begin{cases}1,&x>0\\0,&x<0\end{cases}$ & Kaiming normal\cite{he2015delving} \\
			\hline
			ELU~\cite{clevert2015fast} & $f(x)=\begin{cases}x,&x>0\\ \alpha(e^x-1),&x\leq0\end{cases}$ & $f'(x)=\begin{cases}1, &x>0\\ \alpha e^x, &x<0\end{cases}$ & Normal \\
			\hline
			SELU~\cite{klambauer2017self} & $f(x)=\begin{cases}\lambda x, &x>0\\\lambda\alpha( e^x-1),&x\leq0\end{cases}$ & $f'(x)=\begin{cases}\lambda,&x>0\\\lambda\alpha e^x&x\leq0\end{cases}$ & Normal  \\
			\hline
			Sigmoid & $f(x)=\frac{1}{1+e^{-x}}$ & $f'(x)=f(x)(1-f(x))$ & Xavier normal~\cite{glorot2010understanding}  \\
			\hline
			Softplus & $f(x)=ln(1+e^{x})$ & $f'(x)=\frac{1}{1+e^{-x}}$ & Kaiming normal\cite{he2015delving}  \\
			\hline
			Tanh & $f(x)=\frac{e^x-e^{-x}}{e^x+e^{-x}}$ & $f'(x)=1-(f(x))^2$ & Xavier normal~\cite{glorot2010understanding}  \\
			\hline
			Sine~\cite{sitzmann2020implicit} & $f(x) = sin(x)$ & $f'(x)=cos(x)$ & Specific uniform~\cite{sitzmann2020implicit}  \\
			\hline
			\hline
	\end{tabular}}
\end{table}

\begin{table}[t]
	\setlength\tabcolsep{4pt}
	\centering
	\caption{Quantitative comparisons of different activation functions on the task of (grayscale) image regression.}
	\label{tab:grayreg}
	\begin{tabular}{c|c|c}
		\hline\hline
		\textbf{Activation} & \textbf{PSNR}$\uparrow$ & \textbf{SSIM}$\uparrow$ \\
		
		\hline
		% Abs & 23.48 & 0.655 \\
		% \hline
		% Abs P.E. & 27.81 & 0.775 \\
		% \hline
		ReLU & 25.06 & 0.711 \\
		\hline
		ReLU P.E. & 29.36 & 0.856 \\
		\hline
		ELU & 21.13 & 0.537 \\
		\hline
		ELU P.E. & 29.76 & 0.879 \\
		\hline
		SELU & 20.20 & 0.433 \\
		\hline
		SELU P.E. & 28.55 & 0.837 \\
		\hline
		Sigmoid & 18.06 & 0.408 \\
		\hline
		Sigmoid P.E. & 24.27 & 0.638 \\
		\hline
		Softplus & 18.38 & 0.417 \\
		\hline
		Softplus P.E. & 24.24 & 0.639 \\
		\hline
		Tanh & 23.27 & 0.635 \\
		\hline
		Tanh P.E. & 31.14 & 0.902 \\
		\hline
		Sine & \textbf{33.13} & \textbf{0.960} \\
		
		\hline
		\hline
	\end{tabular}
\end{table}

\section{Precision of Approximate Image Derivatives}\label{sec:precision}

In all aforementioned experiments, the partial derivatives we leverage for supervision are obtained by applying Sobel filters on the images before downsampling and for training, we perform nearest-neighbor downsampling for both image values and approximated derivatives. This is generally reasonable since in real applications, the original images are usually of very high resolution, such as LLFF dataset~\cite{mildenhall2019local,mildenhall2020nerf}, whose original resolution is $3024\times 4032$. Without getting deteriorated rendering results, we would like the training data of INRs to be as few as possible for faster training and a direct solution is to downscale the images. 
As our results in the main paper show, by keeping the derivatives obtained at a high resolution, the problem of downgraded performance will get alleviated.

We also conduct experiments of image regression and inverse rendering where derivatives are obtained from the downsampled images, abandoning the data dependence of the raw data before downsampling, which is to say, we will use all samples in hand for training.

\subsection{Image Regression}
The dataset we used consists of 5 images of $1356\times2040$, which are generated by uniformly sampling images from DIV2K~\cite{Agustsson_2017_CVPR_Workshops} validation set. 
We downsample the original images by nearest-neighbor interpolation with a factor of 4 and use Sobel filters to calculate the approximate derivatives on downsampled images. The network architecture and training \& evaluation settings are consistent with the image regression task of the main paper.

\subsubsection{Results}
The quantitative results are shown in \cref{tab:div2k}. From \cref{tab:div2k}, we can see training with derivatives from downsampled images still gives a great performance improvement than value-based training paradigm.

\subsection{Inverse Rendering}

The experiment settings are different from the main paper. Here we do not perform downsampling to images; instead, we use images of $756\times 1008$ for training and render novel views of $756\times 1008$ and $3024\times 4032$. The approximated image derivatives are calculated from images of $756\times 1008$. As mentioned earlier, LLFF dataset~\cite{mildenhall2019local} provides raw images of $3024\times 4032$ so we can evaluate rendering results at a higher resolution.

\subsubsection{Results}
The quantitative results on rendering views of different resolutions are shown in \cref{tab:supp_ratio}. As can be proven by the results, we can still get a slight performance improvement when training with approximated derivatives from downsampled training images. Also, the performance gap is enlarged, when the rendering views' resolutions are higher, indicating the great generalizability of Sobolev trained MLPs.

The difference between the two sources of image derivatives is basically the precision of approximated derivatives. As shown in the results of both tasks, the precision of approximated derivatives does not affect much. It is using the derivatives for supervision that matters.

\begin{table}[t]
	\setlength\tabcolsep{4pt}
	\centering
	\caption{Quantitative results on the task of image regression on DIV2K validation set.}
	\label{tab:div2k}
	\begin{tabular}{c|c|ccccc}
		\hline\hline
		\multirow{2}*{\textbf{Method}} & \multirow{2}*{\textbf{Mean}} & \multicolumn{5}{c}{DIV2K Validation Set}\\
		\cline{3-7}
		& & \textit{0820} & \textit{0840} & \textit{0860} & \textit{0880} & \textit{0900}\\
		\hline
		\multicolumn{7}{c}{PSNR $\uparrow$}\\
		\hline
		ReLU+P.E.~\cite{mildenhall2020nerf} &  23.34 & 20.44 & 25.75 & 18.62 & 29.36 & 22.51\\ 
		\hline
		ReLU+P.E.+S.T. & 23.91 & 20.90 & 26.30 & 19.17 & 30.00 & 23.18 \\
		\hline
		SIREN~\cite{sitzmann2020implicit} & 22.69 & 19.70 & 24.87 & 17.23 & 29.21 & 22.45 \\
		\hline
		SIREN+S.T. & \textbf{24.44} & \textbf{21.58} & \textbf{26.87} & \textbf{19.35} & \textbf{30.64} & \textbf{23.74}\\
		
		\hline\hline
		\multicolumn{7}{c}{SSIM $\uparrow$}\\
		\hline
		ReLU+P.E.~\cite{mildenhall2020nerf} & 0.577 & 0.497 & 0.651 & 0.371 & 0.817 & 0.547\\ 
		\hline
		ReLU+P.E.+S.T. & 0.625 & 0.559 & 0.692 & 0.401 & 0.843 & 0.629 \\
		\hline
		SIREN~\cite{sitzmann2020implicit} & 0.594 & 0.508 & 0.650 & 0.356 & 0.842 & 0.616\\
		\hline
		SIREN+S.T. & \textbf{0.703} & \textbf{0.663} & \textbf{0.754} & \textbf{0.488} & \textbf{0.875} & \textbf{0.733}\\
		
		\hline\hline
	\end{tabular}
\end{table}

\begin{table}[t]
	\centering
	\caption{Quantitative results of using different paradigms when training with images of resolution $756\times1008$ and rendering novel views of $756\times1008$ and $3024\times4032$.}
	\label{tab:supp_ratio}
	\resizebox{\linewidth}{!}{
		\begin{tabular}{c|c|c|cccccccc}
			\hline\hline
			\textbf{Method} & Render $H\times W$ & \textbf{Mean} & \textit{Fern} & \textit{Flower} & \textit{Fortress} & \textit{Horns} & \textit{Leaves} & \textit{Orchids} & \textit{Room} & \textit{T-Rex}\\
			\hline
			\multicolumn{10}{c}{PSNR $\uparrow$}\\
			\hline
			\multirow{2}*{ReLU+P.E.~\cite{mildenhall2020nerf}} & $756\times 1008$ & 24.69 & 24.26 & 25.92 & 28.58 & 25.32 & 20.39 & \textbf{19.991} & \textbf{28.54} & 24.52\\
			\cline{2-11}
			& $3024\times 4032$ & 23.12 & 22.24 & 24.93 & 27.22 & 23.66 & 18.90 & 19.258 & \textbf{26.24} & 22.54 \\
			\hline
			\multirow{2}*{SIREN+P.E.+S.T.} & $756\times 1008$ & \textbf{24.72} & \textbf{24.33} & \textbf{26.06} & \textbf{28.66} & \textbf{25.40} & \textbf{20.65} &  19.985 & 28.03 & \textbf{24.68} \\
			\cline{2-11}
			& $3024\times 4032$ & \textbf{23.22} & \textbf{22.32} & \textbf{25.07} & \textbf{27.26} & \textbf{23.72} & \textbf{19.10} & \textbf{19.261} & 26.14 & \textbf{22.88} \\
			\hline\hline
			\multicolumn{10}{c}{SSIM $\uparrow$}\\
			\hline
			\multirow{2}*{ReLU+P.E.~\cite{mildenhall2020nerf}} & $756\times 1008$ & 0.755 & 0.744 & 0.788 & 0.792 & 0.746 & 0.646 & 0.606 & \textbf{0.900} & 0.816\\
			\cline{2-11}
			& $3024\times 4032$ & 0.723 & 0.714 & 0.773 & 0.819 & 0.703 & 0.570 & 0.616 & \textbf{0.860} & 0.732 \\
			\hline
			\multirow{2}*{SIREN+P.E.+S.T.} & $756\times 1008$ & \textbf{0.767} & \textbf{0.755} & \textbf{0.807} & \textbf{0.819} & \textbf{0.760} & \textbf{0.666} & \textbf{0.614} & 0.899 & \textbf{0.818}\\
			\cline{2-11}
			& $3024\times 4032$ & \textbf{0.729} & \textbf{0.717} & \textbf{0.781} & \textbf{0.830} & \textbf{0.705} & \textbf{0.581} & \textbf{0.622} & 0.859 & \textbf{0.736} \\
			\hline\hline
	\end{tabular}}
\end{table}

\section{Use of Existing Assets}
Some codes of image regression task and audio regression task are borrowed from \href{https://github.com/vsitzmann/siren}{SIREN}~\cite{sitzmann2020implicit}.
The implementation of inverse rendering task are based on \href{https://github.com/yenchenlin/nerf-pytorch}{NeRF-PyTorch}~\cite{lin2020nerfpytorch}, which is a PyTorch version of original \href{https://github.com/bmild/nerf}{NeRF}~\cite{mildenhall2020nerf}.

%\begin{figure}[t]
%	\centering 
%	\subfigure[SIREN~\cite{sitzmann2020implicit}]{
%		\includegraphics[width=0.8\linewidth]{camera_ready/figures_supp/bach_sine.pdf}}
%	\subfigure[SIREN+S.T.]{
%		\includegraphics[width=0.8\linewidth]{camera_ready/figures_supp/bach_sine_grad.pdf}}
%	\caption{Regressed waveforms of the INRs trained with and without derivative supervision on the clip of \textit{Bach}. From top to bottom: ground truth waveform, regressed waveform and error detected. We zoom in the waveform marked with red at the right column accordingly.} 
%	\label{fig:bach}
%\end{figure}

%\begin{figure}[t]
%	\centering 
%	\subfigure[SIREN~\cite{sitzmann2020implicit}]{
%		\includegraphics[width=0.8\linewidth]{camera_ready/figures_supp/counting_sine.pdf}} 
%	\subfigure[SIREN+S.T.]{
%		\includegraphics[width=0.8\linewidth]{camera_ready/figures_supp/counting_sine_grad.pdf}} 
%	\caption{Regressed waveforms of the INRs trained with and without derivative supervision on the clip of \textit{Counting}. From top to bottom: ground truth waveform, regressed waveform and error detected. We zoom in the waveform marked with red at the right column accordingly.} 
%	\label{fig:counting}
%\end{figure}

\end{document}